%% file: output.tex
\documentclass[10pt,journal,compsoc]{IEEEtran}

\newcommand{\etal}{\textit{et al.}}
\usepackage{pifont}
\newcommand{\cmark}{\ding{51}}%
\newcommand{\xmark}{\ding{55}}%

\usepackage{graphicx}
\usepackage{subcaption}
\usepackage{multirow}
\usepackage{amsmath}
\usepackage{hyperref}
\usepackage{xcolor,colortbl}
\usepackage{comment}
\usepackage{amssymb}
\usepackage{diagbox}
\usepackage{makecell}
\newcolumntype{x}[1]{>{\centering\let\newline\\\arraybackslash\hspace{0pt}}p{#1}}

\usepackage{enumitem}

\ifCLASSINFOpdf
 
\else

\fi

\begin{document}

\title{SiT: Self-supervised vIsion Transformer}

\author{Sara~Atito,~\IEEEmembership{Member~IEEE,}
        Muhammad~Awais,~\IEEEmembership{Member~IEEE,}
        and~Josef~Kittler,~\IEEEmembership{Life Member,~IEEE}
\IEEEcompsocitemizethanks{\IEEEcompsocthanksitem Centre for Vision, Speech and Signal Processing (CVSSP), University of Surrey, Guildford, United Kingdom
\IEEEcompsocthanksitem {\texttt \{s.a.ahmed,muhammad.awais,j.kittler\}@surrey.ac.uk}}
\thanks{Manuscript received ?? ??, 2021; revised ?? ??, 2021.}}

\markboth{Journal of \LaTeX\ Class Files,~Vol.~??, No.~??, ??~??}%
{Atito \MakeLowercase{\textit{et al.}}: SiT: Self-supervised vIsion Transformer}
\IEEEtitleabstractindextext{%
\begin{abstract}
In Natural Language Processing (NLP), Self-supervised Learning (SSL) and transformers are already the methods of choice due to the tremendous success of attention based self-supervised transformer models like BERT~\cite{devlin2018bert} and GPT~\cite{radford2018improving}. 
So far, the vision transformers, adopted from NLP transformers, have been shown to work well when pretrained either using a large scale supervised data~\cite{dosovitskiy2020image} or with some kind of co-supervision, e.g. in terms of teacher network~\cite{touvron2020training}. These supervised pretrained vision transformers achieve outstanding results in downstream tasks with minimal changes~\cite{dosovitskiy2020image,touvron2020training,el2021training}.  Self-supervised Pretraining (SSP) is still not the method of choice for computer vision due to performance gap~\cite{dosovitskiy2020image}, however, SSL is gaining increasing traction in computer vision as the performance gap between Supervised Pretraining (SP) and SSP is reducing for downstream applications, like classification, localisation, segmentation, etc. 
{\bf S}elf-supervised v{\bf i}sion {\bf T}ransformers ({\bf SiT}) is the first work which establishes that SSP can outperform SP for downstream applications, establishing SSP as a more suitable choice for pretraining vision transformers.  

SiT is the first masked image modelling work for vision transformers. At its core SiT builds the idea of Group Masked Model Learning (GMML), a simple masked autoencoder framework to obtain a pretext model. 
The architectural flexibility of vision transformers allows us to use SiT as an autoencoder and work with multiple self-supervised tasks seamlessly.  
The proposed approach is evaluated on standard datasets using common protocols. The results demonstrate the suitability of the GMML framework for SSL and vision transformers. SiT consistently outperforms supervised pretraining as well as prior arts with a large margin. Unlike other vision transformer based pretraining methods, SiT performs very strongly on small and medium scale datasets as well. 
Thanks to SiT, the vision transformers can outperform (perform on par with) Convolutional Neural Network (CNN) counterpart for small and medium datasets without using any external data for pretraining, overcoming the problem of data-hungry vision transformers. 
Pretraining, finetuning, and evaluation codes are available under: \href{https://github.com/Sara-Ahmed/SiT}{https://github.com/Sara-Ahmed/SiT}. 

\textbf{Impact:} We proposed GMML framework in SiT for self-supervised learning of vision transformers at the beginning of 2021 using masked autoencoder with reconstruction loss, however the idea is generally applicable to other losses as shown in later studies~\cite{bao2021beit,atito2021mcssl,zhou2021ibot}. At the time of conception of SiT, the merits of GMML were shown employing small models and small/medium scale datasets due to extremely restricted computational resources. Since then, GMML has been widely adopted in computer vision and other related fields. Towards the end of 2021, SIMMIM~\cite{xie2021simmim} and MAE~\cite{he2021masked} extended GMML with reconstruction loss using huge vision transformers on large scale datasets, like ImageNet-1K~\cite{deng2009imagenet}. GMML is now the leading SSL framework on multiple application areas, giving sate-of-the-art results for image classification~\cite{atito2021mcssl}, segmentation~\cite{xie2021simmim}, audio analysis~\cite{gong2021ssast}, medical image analysis~\cite{zhou2022self,chen2022masked}, video representation~\cite{tong2022videomae}, etc. In short MIM/GMML is enabling the computer vision community to enjoy the same success in SSL which NLP community has enjoyed for BERT. 
SiT performs much better than prior art and post art when trained using small to medium scale dataset without any external data and performs better than prior art and on par with post art which have adopted GMML framework when pretrained on large scale datasets. 
\end{abstract}

\begin{IEEEkeywords}
Masked Image Modelling (MIM), Masked autoencoders, Group Masked Model Learning (GMML), Vision Transformer, Self-supervised Learning, Discriminative Learning, Image Classification, Transformer-based Autoencoders.
\end{IEEEkeywords}}

\maketitle

\IEEEdisplaynontitleabstractindextext

\IEEEpeerreviewmaketitle

\IEEEraisesectionheading{\section{Introduction}
\label{sec:intro}}

\IEEEPARstart{R}{ecent} trends, particularly in NLP,  showed that self-supervised pretraining can improve the performance of downstream tasks significantly \cite{devlin2018bert,brown2020language}. Similar trends have been observed in speech recognition \cite{baevski2020wav2vec} and computer vision applications \cite{he2020momentum,chen2020simple,grill2020bootstrap,caron2020unsupervised}. 
The self-supervised pretraining, particularly in conjunction with transformers~\cite{vaswani2017attention}, 
are the models of choice for NLP~\cite{devlin2018bert,brown2020language}. The success of SSL comes at the cost of massive datasets and huge capacity models. For instance, NLP based transformers are trained on hundreds of billions of words consisting of models with several billion parameters \cite{brown2020language}. 
The recent success of Transformers in image classification~\cite{dosovitskiy2020image} generated a lot of interest in the computer vision community. However, the pretraining of vision transformer mainly thrive using very large scale datasets using supervised learning, e.g., datasets consisting of hundred of millions of labelled samples~\cite{dosovitskiy2020image}. This particular data-hungry nature of vision transformers arise due to lack of so called inductive bias~\cite{atito2021mcssl}. 
Very recently vision transformer have been shown to perform well on ImageNet-1K without external data~\cite{touvron2020training}. However, they need distillation approaches and guidance from CNNs counterparts. In short, pretraining a neural network using large-scale supervised datasets is a norm in computer vision in order to obtain better performance. However, the manual annotation of training data is quite expensive, despite the engineering innovations of crowd sourcing campaigns. More importantly development of the visual cortex and visual memory seems to depends upon the visual experience~\cite{hubel1970period,blakemore1970development}. This is suggested by the early plasticity experiments on kittens~\cite{hubel1970period,blakemore1970development} which support the argument that some important aspect of visual perception are acquired through learning and visual experiences. 
Training of DNNs via supervised learning with one label per image may corresponds to having limited visual experience for trained DNNs because DNNs may not learn from rich visual information present in other concepts in the natural images. This may effect the generalisation capability of the DNNs. 
Furthermore, learning using labels as a supervisory signal, particularly one label per natural image, can be though of as an ill-posed problem. 
The DNNs may map an input image to a target class and in the process have to be invariant to other concepts. In natural images there could be multiple concepts common between different images while the single annotated label could be different among them. This may cause confusion for the DNN and may result in sub-expressive features from labelled data. Labelling every salient concept in every images be also be infeasible. 
To address these limitation, SSL methods \cite{schmidhuber1987evolutionary,wang2015unsupervised,pinto2016curious,he2020momentum,grill2020bootstrap,caron2020unsupervised,chen2020generative} have been proposed to train more generalisable DNNs suitable for several downstream tasks and construct image representations that are semantically meaningful from unlabelled data. 

Self-supervised methods can roughly be categorised in to generative and discriminative approaches. Generative approaches \cite{doersch2015unsupervised,dumoulin2016adversarially,NEURIPS2019_18cdf49e} learn to model the distribution of the data. However, data modelling generally is computationally expensive and may not be necessary for representation learning in all scenarios. On the other hand, discriminative approaches, typically implemented in a contrastive learning framework \cite{caron2018deep,hjelm2018learning,chen2020simple,NEURIPS2020_29539ed9} or using pre-text tasks \cite{larsson2017colorization,gidaris2018unsupervised,jenni2018self}, demonstrate the ability to obtain better generalised representations with modest computational requirements.

The primary focus of contrastive learning is to learn image embeddings that are invariant to different augmented views of the same image while being discriminative among different images. Despite the impressive results achieved by contrastive learning methods, they often disregard the learning of contextual representations as they are focusing on one global transformation invariant representation for the whole image. While each and every concept in the image and context within that concept and context around that concept is important for in depth understanding of the image. Moreover, most contrastive learning approaches suffer from collapse, a trivial constant solutions. To avoid the collapse these, methods use careful implementation details, e.g. stop gradients, large batch size, exponential moving average based teacher network, centring, asymmetric projection head, etc. 
For more detail and context aware representations, alternative pretext tasks, such as reconstruction or recovery of missing information based approaches, might be better suited. 
In recent years, a stream of novel pretext tasks have been proposed in the literature, including inpainting patches \cite{pathak2016context}, colourisation \cite{zhang2016colorful,larsson2016learning,larsson2017colorization}, relative patch location \cite{doersch2015unsupervised}, solving jigsaw puzzles \cite{noroozi2016unsupervised,kim2018learning}, cross-channel prediction \cite{zhang2017split}, predicting noise \cite{bojanowski2017unsupervised}, predicting image rotations \cite{gidaris2018unsupervised}, spotting  artefacts \cite{jenni2018self}, etc. These pretext tasks have been explored for SSL using CNNs frameworks. Different from them we developed pretext framework for Vision Transformers (ViT) which can capture local and global context seamlessly. Unlike CNNs transformers do not make any assumption about local inductive bias (statistical correlation in the neighbourhood), hence, in order to model useful inductive bias, ViTs require huge amount of data to perform on par with CNNs. The proposed GMML framework enables ViTs to learn the useful local inductive bias even from small amount of data and enables ViTs to perform on par with CNNs even on small data while maintaining the advantage on large data.

The core of SiT is built upon the simple idea of GMML. Different from existing SSL approaches, GMML leverage the information redundancy and complementarity in the vision transformers by learning to recover/reconstruct local content by linking it to context. In spirit, this principle is similar to the masked language modelling (MLM) used in BERT~\cite{devlin2018bert} which recover masked words from context. The principle is also inspired from word2vec~\cite{mikolov2013distributed} which predict words from the context. 
In computer vision, we take the inspiration from the principle of denoising autoencoder~\cite{vincent2008extracting} and from the idea of context encoder~\cite{pathak2016context} which has been studied for unsupervised learning using CNNs. 
The GMML extends the principles of MLM, denoising autoencoders, and context encoders to vision transformers for self-supervised learning. 
This is achieved by three principles: i) {\it learning to recover the input stimulus by a mechanism akin to autoencoding, implemented by means of random data tokens perturbation using masking of groups of connected tokens, etc.}
ii) {\it a perception-action mechanism} \cite{Granlund-ivc09}, {\it which learns to recognise an action from its impact on perception}, and iii) {\it learning the notion of similarity of content from the preservation of content identity in the data.} The proposed SSL approach is instrumental in extracting an intrinsic data model and is admirably able to adapt to downstream tasks by finetuning. The GMML establishes itself as a  strong standalone SSL framework surpassing all existing SSL methods and additionally outperforming supervised pretraining for the first time. Thanks to architectural flexibility of transformers, SiT further extend GMML and leverages the advantages of both contrastive learning and pre-text approaches. 

The main contributions of this study are summarised as follows:
\begin{enumerate}[leftmargin=*]
\item We propose Group Masked Model Learning (GMML), a novel framework for self-supervised learning of visual representations using vision transformers. GMML trains DNNs and learns rich representations by recovering large amount (upto 70\%) of missing visual information by groups of masked tokens using the context present in the visible tokens. 
\item We endow the GMML architecture with a decoder and demonstrate that it can be implemented by essentially using a 2-layer perceptron, thanks to the intrinsic characteristics of the transformer. This transformer based autoencoder avoids the need for a whole decoder block which is typically present in CNNs based encoder-decoder architectures.
\item Drawing on the natural ability of the autoencoding transformer to support multi-task learning, we develop a strong self-supervised framework which jointly optimises the reconstruction (GMML) and contrastive losses.   
\item We illustrate the effectiveness of the proposed framework on standard benchmarks following different evaluation protocols including domain transfer and finetuning.
\item We outperform the concurrent and post arts in different datasets with a large margin reaching +5.4\% improvements when the models are pretrained on small datasets and obtain on par performance with the state-of-the-art when the models are pretrained on large-scale datasets.
\end{enumerate}
There are three key observations summarised as follow:
\begin{enumerate}[leftmargin=*]
\item The ability of SiT in clustering the data without any form of supervision.
\item With Sit, it is possible to train data hungry transformers on tiny datasets with just a few thousand sample such as Flowers, Pets, CIFAR10, etc without distillation. 
\item Most importantly to best of our knowledge at the beginning of 2021 SiT became the first work showing that self-supervised pretraining can consistently outperforms supervised pretraining for the vision classification downstream tasks using transformers.

\end{enumerate}

\noindent
The paper is structured as follows. Section \ref{sec:relatedworks} provides a background on the state-of-the-art self-supervised techniques. In Section \ref{sec:method}, the proposed self-supervised framework using vision transformer is explained. The experimental analysis and a discussion of the obtained results are shown in Section \ref{sec:exp}. Finally, conclusions of this study are presented in Section \ref{sec:conc}.

\section{Related Works}
\label{sec:relatedworks}

\subsection{Comparison with Prior Art}

Discriminative approaches to SSL~\cite{caron2018deep,hjelm2018learning,chen2020simple,NEURIPS2020_29539ed9} typically have been demonstrated to learn better representations than generative approaches~\cite{doersch2015unsupervised,dumoulin2016adversarially,NEURIPS2019_18cdf49e} and hence will be focus on the literature review.
Discriminative approaches are typically implemented using pre-text tasks or in a contrastive learning framework. The basic pretraining mechanism of the handcrafted pre-text tasks is autoencoding~\cite{kramer1991nonlinear}, which forces a network to find a representation that allows the reconstruction of the input image, even if corrupted by perturbations or noise.
Many self-supervised pretext tasks manipulate the input data to obtain better image representations. For example, Pathak~\etal~\cite{pathak2016context} trained a convolutional network to predict the content of arbitrary missing regions in an image based on the rest of the image. The motivation behind this work is that for the network to produce plausible hypothesis for the missing parts, the encoder needs to understand the content of the entire image. In the same line, Zhang \etal \cite{zhang2016colorful} proposed an image colourisation task by predicting a coloured version of the given grey-scale input image and used class re-balancing to increase the diversity of the predicted colours. 
Furthermore, Doersch~\etal~\cite{doersch2015unsupervised} presented one of the pioneering works of using spatial context information for feature learning by training a convolutional network to recognise the relative positions of random pairs of image patches. 
Following this idea, several methods were proposed to learn image features by solving even more difficult spatial context tasks (e.g. jigsaw puzzles \cite{noroozi2016unsupervised,kim2018learning}). 
Training the network with such objective by using the within-image context encourages the network to learn local feature representations while ignore global context. 
Gidaris~\etal~\cite{gidaris2018unsupervised} proposed RotNet, a convolutional network that learns image features by training the network to recognise a pre-defined 2d rotation that is applied to the input image. Using this simple task DNNs can learn global image level visual representations based on the assumption that network will have understanding of the object if it predict the objects' orientation. 
Overall, such pretext based approaches are powerful in learning useful representations from unlabeled data, yet, they limit the generality of learning discriminative representations between different samples, where contrastive approaches are better suited. 

Contrastive approaches~\cite{oord2018representation,wu2018unsupervised,hjelm2018learning,bachman2019learning,
chen2020simple,NEURIPS2020_29539ed9,grill2020bootstrap} train the network by bringing the representations of different augmented views of the same image closer and spreading the representations of views from different images apart. In general contrastive learning based approaches tend to perform better than pretext task based approaches. 
Chen \etal \cite{chen2020simple} proposed SimCLR, a contrastive self-supervised learning algorithm without requiring specialised architectures or a memory bank. SimCLR is a simple framework to learn representations from unlabeled images based on heavy data augmentation by maximising the similarity between two augmented views coming from the same image. Training the network with such objective improves the quality of the learnt representations in discriminating between samples drastically. Contrastive learning approaches either use large batch sizes~\cite{chen2020simple} or memory banks~\cite{wu2018unsupervised,he2020momentum} in order to have informative negative samples in the batch. These approaches typically use various tricks to avoid representation collapse. 

Deep clustering-based methods~\cite{caron2018deep,haeusser2018associative,asano2019self,caron2020unsupervised,li2020prototypical} learn representation by clustering the images in the embedding space. 
DeepCluster~\cite{caron2018deep} clusters data points using the current representation 
to produce labels for the next representation. The cluster index of each sample is then used as a classification target for the new representation. This approach is computationally expensive as it requires a clustering phase with precautions to avoid collapsing to trivial solutions.

Hjelm \etal \cite{hjelm2018learning} investigated the use of mutual information for unsupervised representation learning through Deep InfoMax by maximising the mutual information in global and local scales across structural patches in an image following the InfoMax principle \cite{linsker1988self}.

Patacchiola and Storkey \cite{NEURIPS2020_29539ed9} proposed a self-supervised formulation of relational reasoning that allows a learner to bootstrap a signal from the information implicit in unlabelled data. Specifically, they used a relation network as a learnable function on the unlabeled dataset to quantify the relationships between augmented views of the same object (i.e. intra-reasoning) and between different objects in different scenes (i.e. inter-reasoning) which could help learners to distinguish the object based on their differences.

In this work, we leverage the advantage of both pre-text approaches and contrastive learning approaches to learn useful as well as discriminative representations between different samples employing a simple transformer-based framework for self-supervised learning.

\begin{figure*}[!t]
    \centering
    \includegraphics[width=0.9\linewidth]{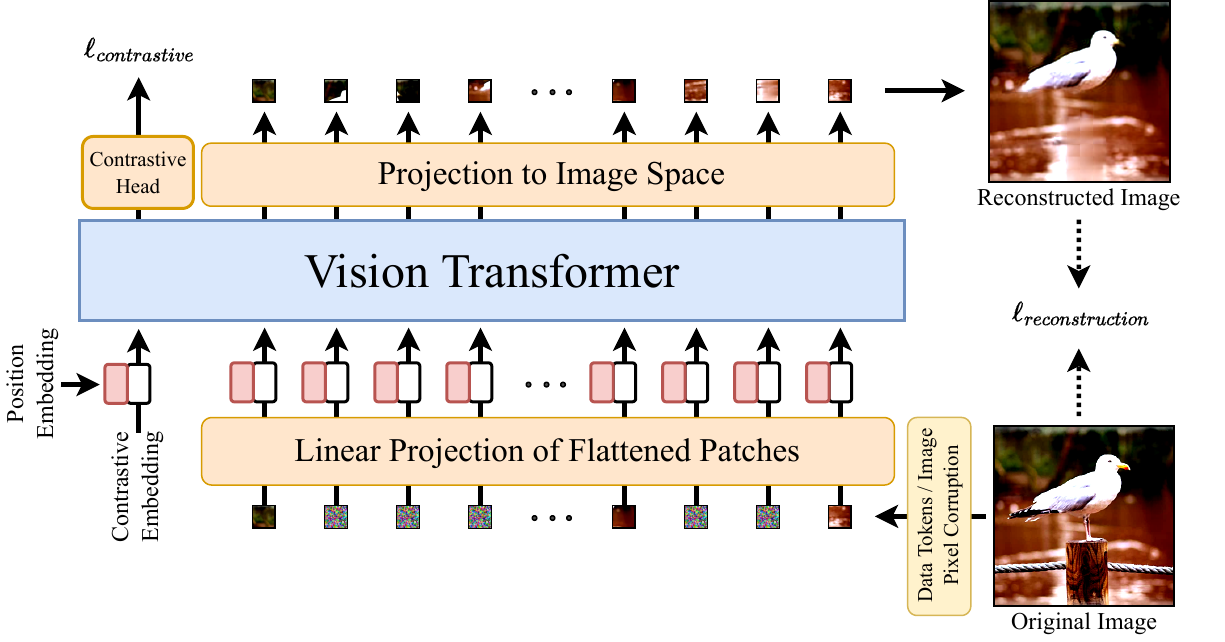}
    \caption{Self-supervised vIsion Transformer (SiT) }
    \label{fig:transformer}
\end{figure*}

\subsection{Comparison with Post Art}

Recently, a manifold of methods have used the principals outlined in GMML at the beginning of 2021. In this section, we will briefly introduce the similarities and differences between GMML and some of the most popular post art. 

Two notable post arts are SimMIM~\cite{xie2022simmim} and MAE~\cite{Kaiming2021mae}. Similar to GMML, both SimMIM and MAE use the principal of transformer based masked autoencoder. Both of them mask a high proportion of data-tokens randomly. However, we note that masking a very high proportion of the data-token essentially defines groups of connected tokens to be masked. We also note that  
their optimal masking proportion is very similar to GMML. 
Following DropToken idea in VATT~\cite{akbari2021vatt}, MAE discard the masked tokens for encoder and use them in decoder to reconstruct the image.
However, this dropping of masked tokens require MAE to use complex decoder consisting of six to twelve layers of transformers, unlike GMML which use 2 pointwise convolutional layers. We noticed that wall clock time for the pretraining of MAE and GMML is similar for ViT-B, while training time for ViT-S is much slower for MAE as compared to GMML due to complex decoder of MAE. Furthermore, due to lack of modelling the inductive bias, the
performance of MAE degrade largely for small datasets and MAE only performs on par with GMML for large dataset. SimMIM is very similar to GMML the only meaningful difference is that GMML uses noise and alien concepts in addition of masking with zero while SimMIM just uses masking with zeros. Besides, the corruption in SimMIM is applied after the patch projection block whilst in GMML, the corruption is applied directly to the image pixels.  

Another noticeable method in post art is BeIT~\cite{bao2021beit}. BeIT uses external knowledge by using an encoder trained without supervision, to group visual patches in order to define a visual vocabulary. This enables the use of cross entropy as a loss function, like in BERT~\cite{devlin2018bert}. However, unlike BERT the classes are coming from external knowledge source albeit trained unsupervisedly. It can be considered as an expensive and extreme case of patch level distillation via supervised or unsupervised encoder. 
Secondly, it will inherit issues of visual vocabulary, like, a fixed number of visual words, a quantisation error, visual ambiguity when assigning to cluster centres etc. 

\section{Methodology}
\label{sec:method}
Supervised learning, as demonstrated in \cite{dosovitskiy2020image}, allows the transformer to learn a bottleneck representation where the mixing of content and context is centred primarily about the class token. This creates a rather superficial model of the data, and its linking to labels requires a huge number of samples for training.
In contrast, GMML based unsupervised learning exploits information redundancy and complementarity in the image data by learning to reconstruct local content by integrating it with context. 
The proposed self-supervised learning approach is instrumental in extracting an intrinsic data model, that is robust to perturbations and is admirably able to adapt to downstream tasks by finetuning.  The proposed approach offers remarkable advantages: 
\begin{itemize}
\item The self-supervised transformer can be trained with unlabelled data. 
\item The amount of labelled training data required for finetuning to learn a downstream task is two orders of magnitude lower than the counterpart needed for direct training. 
\item The total amount of training data (labelled and unlabelled) is also several orders of magnitude lower.
\item The performance achieved is significantly better than state-of-the-art self-supervised methods.
\end{itemize}
The proposed methodology of transformer pretraining by self-supervision is expected to have a significant impact on the advancement of science by enabling the wider research community starved of resources to contribute to deep learning. 

Thus the main goal of this work is to learn a representation of the data in an unsupervised fashion. This is achieved by recovering partially masked or transformed local parts of the image represented by data-tokens at the input of the vision transformer. The underlying hypothesis is that, by recovering the corrupted tokens/parts of an image from the uncorrupted tokens/part based on  the context from the whole visual field, the network will implicitly learn the notion of visual integrity. This notion of visual integrity is further enhanced by using pseudo labels that can be generated automatically based on some attributes of the data. Learning from recovery of the transformed parts and learning from pseudo label may seem different but the underlying motivation behind both kinds of self-supervised learning mechanisms is the same, i.e., learning visual integrity. For example, intuitively the network will only be able to recover the pseudo labels if it learns the characteristic properties of visual stimuli corresponding to specific actions impacting on the visual input.  The weights of the learned model can then be employed as an initialisation point for any downstream task like image classification, object detection, segmentation, etc. To achieve this goal, we propose a Self-supervised vIsion Transformer (SiT) in which the model is trained via group masked model learning and to estimate different geometric transformations applied to the input image and hence, better image representation can be obtained. 

\subsection{Self-Supervised Vision Transformer}
\label{subsec:ss-tiv}

Transformer~\cite{vaswani2017attention} has shown great success in various natural language processing tasks~\cite{devlin2018bert,radford2019language,brown2020language}. Recently, many researchers attempted to explore the benefits of transformer-based models in computer vision tasks \cite{dosovitskiy2020image,touvron2020training}. In this work, we introduce Self-supervised vIsion Transformer (SiT) adapted from Vision Transformer (ViT) proposed by \cite{dosovitskiy2020image} with some modifications. 

Vision Transformer~\cite{dosovitskiy2020image} receives as input a sequence of patches obtained by tokenizing the input image $\mathbf{x} \in \mathbb{R}^{H \times W \times C}$ into $n$ flattened tow-dimensional 
patches of size $p_1 \times p_2 \times C$ pixels, where $H$, $W$, and $C$ are the height, width, and number of channels of the input image, ($p_1\times p_2$) is the patch size, and $n$ is the number of patches, i.e. $n = \frac{H}{p_1} \times \frac{W}{p_2}$. Each patch is then projected with a linear layer to $D$ hidden dimensions. The whole operation can also be implemented simply by a convolutional layer with kernel size $p_1 \times p_2$, number of input and output channels $C$ and $D$ respectively. 
In order to retain the relative spatial relation between the patches, fixed or learnable position embeddings are added to the patch embeddings as an input to the Transformer encoder.

The Transformer encoder consists of $L$ consecutive Multi-head Self-Attention (MSA) and Multi-Layer Perceptron (MLP) blocks. The MSA layer is defined by $h$ self-attention heads where each head outputs a sequence of size $n \times d$. 
The employed self attention mechanism is based on a trainable triplet of (query, key, value). Each query vector in $\mathbf{Q} \in \mathbb{R}^{n\times d}$ is matched against a set of key vectors $\mathbf{K}  \in \mathbb{R}^{n\times d}$, and the output is then normalised with a softmax function and multiplied by a set of values $\mathbf{V} \in \mathbb{R}^{n\times d}$. Thus, the output of the self-attention block is the weighted sum of $\mathbf{V}$ as shown in Equation \ref{eq:SA}. The output sequences of each block are then concatenated into $n \times dh$ and projected by a linear layer into $n \times D$ sequence. 

\begin{equation}
    {\rm SelfAttention}(\mathbf{Q}, \mathbf{K}, \mathbf{V}) = {\rm Softmax}(\frac{\mathbf{Q}\mathbf{K}^T}{\sqrt{d}})\mathbf{V}
    \label{eq:SA}
\end{equation}

To serve the classification task, a trainable vector (i.e class token) is appended to the input sequence of the patch tokens and goes through the Transformer encoder. Finally, a classification head is added to the output of the Transformer encoder corresponding to the class token. The classification head is implemented by a single linear layer that projects the class embeddings to the number of classes.  

Unlike ViT \cite{dosovitskiy2020image}, our work is based on unsupervised learning/pretraining, hence, classification token is not required. Instead we have a contrastive token, beside the data tokens (image patches token) used for image reconstruction, to serve the proposed self-supervised pretraining tasks. The contrastive token adopted from SimCLR~\cite{chen2020simple}, described in Section \ref{subsec:contrastive}, is employed to serve the contrastive prediction task. The main architecture of our proposed network is shown in Figure~\ref{fig:transformer}. 

\subsection{Self-Supervised Tasks}
\label{sec:g_trns}
As noted earlier, the transformer architecture allows seamless integration of multiple task learning simultaneously. We leverage this strength of the transformers to train SiT with two different objectives:
(1) Image reconstruction based on GMML and (2) Contrastive learning. In the rest of the section, we describe the different types of self-supervised tasks employed in this work. 

\subsubsection{Task \#1: Image Reconstruction}

For image reconstruction, we propose to use the transformer as an autoencoder, i.e., visual transformer autoencoder. Unlike CNNs based autoencoders which require encoder and expensive decoders consisting of convolutional and transposed convolution layers, the decoder in the transformer autoencoder can be implemented using a light decoder. One limitation 
of CNN based encoder is the step of information summarisation in which the information is essentially discarded using strided convolutions or max pool operations. This information is then recovered by series of upsampling and convolution operations (or transposed convolutions) with skip connection between encoder and decoder.  By analogy to auto-encoders, our network is trained to reconstruct the input image through the output tokens of the transformer. To learn better semantic representations of the input images, we apply Group Masked Model Learning (GMML) by applying several transformations to local patches of the image. Unlike standard masked tokens in BERT, we apply these local transformations to a block of neighbouring tokens arranged spatially (in 2D rather than in sequence only). In BERT and other NLP based transformers, it makes sense to mask only one token, as a single token can represent semantic meaning. However, for visual signals, it is critical to transform neighbouring tokens consisting of either one or more semantic concepts. The aim is to recover these transformed local parts at the output of SiT. In doing so, SiT implicitly learns the semantic concepts in the image. It should be noted that these transformed tokens can be either on the foreground object or on the background, and recovering these tokens is equally valid for both scenarios. Indeed while modelling the visual signal in this way we are moving away from the notion of foreground and background, and every part of the content is considered as a semantic concept, whether it is a horse grazing in a meadow, or the meadow itself. The intuition is that by modelling all semantic concepts, SiT will generalise better for unseen tasks, whether they are related to an object, a distributed object, or to the whole visual signal. 

Image inpainting is a simple but effective pre-text task for self-supervision, which proceeds by training a network to predict arbitrary transformed regions based on the context. This context can be from the same object on which the transformed region is applied or from the surrounding objects/concepts. With CNNs this context is defined by the, so called, receptive field, while with transformers the context consists of the whole image. The motivation behind image inpainting is that the network  is  required  to  learn  the  knowledge including  the  colour, texture  and  structure  of  the  objects/concepts to infer the missing areas. In this work, we employed two types of image inpainting, random dropping, by randomly replacing neighbouring patches from the image with random noise, and random replacement, by randomly replacing patches from the image with patches from another image. 

The objective of the image reconstruction is to restore the original image from the corrupted image. For this task, we use the $\ell1$-loss between the original and the reconstructed image as shown in Equation \ref{eq:l1-pixel}. Although, $\ell2$-loss generally converges faster than $\ell1$-loss, $\ell2$-loss is prone to over-smooth the edges for image restoration \cite{zhao2016loss}. Therefore, $\ell1$-loss is commonly used for image-to-image processing more than $\ell2$-loss.

\begin{equation}
\label{eq:l1-pixel}
\mathcal{L}_{\rm recons}(\mathbf{W}) = \frac{1}{N}\sum_i^N ||\mathbf{x_i} - {\rm SiT}_{\rm recons}(\bar{\mathbf{x_i}})||
\end{equation}

Where $||.||$ is the $\ell1$ norm, $\mathbf{x_i}$ is the input image, $\bar{\mathbf{x_i}}$ is the corrupted image, $N$ is the batch size, and $\mathbf{W}$ denotes the parameters of the transformer to be learned during training. Finally, ${\rm SiT}_{\rm recons}(.)$ returns the reconstructed image by feeding the output of the data tokens of the backbone $E(.)$, i.e. vision transformer, to the light decoder $D(.)$. Thus, ${\rm SiT}_{\rm recons}(.) = D( E(.)[{\rm data tokens}] )$

\subsubsection{Task \#2: Contrastive learning}
\label{subsec:contrastive}
In self-supervised learning, we do not have any concept labels for the training data. However, by applying geometric and perturbation transformations to a training sample, we do not change the perceptual identity of the content, and the transformer should produce for all such synthetically-generated content-matching pairs a similar output. We adopt the cosine similarity as the underlying measure of similarity of representation. Inspired by recent contrastive learning algorithms \cite{park2020contrastive}, we incorporated a contrastive loss to the objective function where the network is trained to minimise the distance between positive pairs, i.e. augmented images coming from the same input image, and maximise the distance between negative pairs, i.e. samples coming from different input images. In particular, we employed the normalised temperature-scaled softmax similarity \cite{wu2018unsupervised,hjelm2018learning,chen2020simple} between a sample $\mathbf{x_i}$ and any other point $\mathbf{x_j}$ defined as follows:
\begin{equation}
\label{eq:contrastive}
    \ell^{\mathbf{x_i},\mathbf{x_j}}_{\rm contr}(\mathbf{W}) =  \frac{{\rm e}^{{\rm sim}({\rm SiT}_{\rm contr}(\mathbf{x_i}), ~{\rm SiT}_{\rm contr}(\mathbf{x_j}))/\tau}}{\sum_{k=1, k\neq i}^{2N} {\rm e}^{{\rm sim}({\rm SiT}_{\rm contr}(\mathbf{x_i}),~{\rm SiT}_{\rm contr}(\mathbf{x_k}))/\tau}}
\end{equation}

where ${\rm SiT}_{\rm contr}(.)$ denotes the image embedding coming from the contrastive head by feeding the output of the class token of the backbone $E(.)$, to the light contrastive head $Contr(.)$. Thus, ${\rm SiT}_{\rm Contr}(.) = Contr( E(.)[{\rm class token}] )$. As for $\rm sim(.,~.)$, it is the dot product of the $\ell_2$ normalised inputs, i.e. cosine similarity, and $\tau$ denotes a constant temperature parameter which we set to $0.2$ throughout the experiments as suggested by \cite{chen2020simple}.  Rather than using the cosine similarity measure directly, the normalised softmax in (\ref{eq:contrastive}) has the advantage that its optimisation enhances the similarity of matching pairs, as well as the dissimilarity of negative pairs.  Now, let $\mathbf{x_{\tilde j}}$ be a sample matching in content the sample, $x_j$. The contrastive loss, $\mathcal{L}_{\rm contr}(\mathbf{W})$, is then defined as the arithmetic mean over all positive pairs in the batch of the cross entropy of their normalised similarities, i.e.
\begin{equation}
\label{eq:contrastive1}
    \mathcal{L}_{\rm contr}(\mathbf{W}) = -\frac{1}{N}\sum_{j=1}^N \log \ell^{\mathbf{x_j},\mathbf{x_{\tilde j}}}_{\rm contr}(\mathbf{W})
\end{equation}

To perform the contrastive learning, we feed the network the 2 clean and corrupted augmented views from each sample. The loss is then calculated by matching the cross representation between the two views, i.e. matching the representation of the clean first view with the corrupted second view and the representation of the clean second view with the corrupted first view.

There are two ways to obtain the representation of the clean and corrupted views: (1) Feeding both the clean images and the corrupted images to the same encoder (as a siamese network) and stop the back-propagation when clean images are passed to the encoder, (2) Adopting a momentum encoder where the corrupted images are passed to the encoder and the clean images are passed to the momentum encoder. 

In this work, we adopted the second strategy as we found that the performance when employing a momentum encoder is slightly better. For the momentum encoder, we employed the same architecture as the encoder and the weights of the momentum encoder is updated using exponential moving average of the encoder weights.

\begin{figure*}[t]
    \centering

    \begin{subfigure}[t]{0.32\textwidth}
        \centering
        \begin{subfigure}[t]{0.32\textwidth}
        \includegraphics[width=\linewidth]{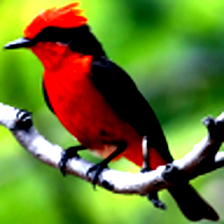}
        \includegraphics[width=\linewidth]{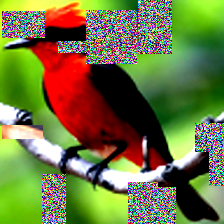}
        \includegraphics[width=\linewidth]{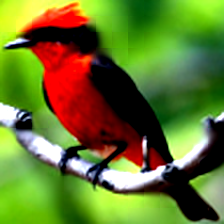}
        \end{subfigure}%
        \hfill
        \begin{subfigure}[t]{0.32\textwidth}
        \includegraphics[width=\linewidth]{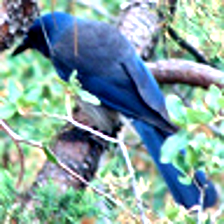}
        \includegraphics[width=\linewidth]{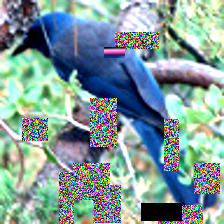}
        \includegraphics[width=\linewidth]{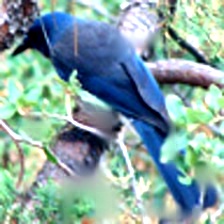}
        \end{subfigure}%
        \hfill
        \begin{subfigure}[t]{0.32\textwidth}
        \includegraphics[width=\linewidth]{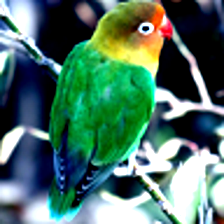}
        \includegraphics[width=\linewidth]{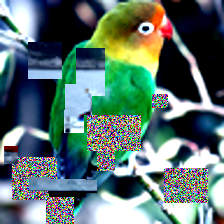}
        \includegraphics[width=\linewidth]{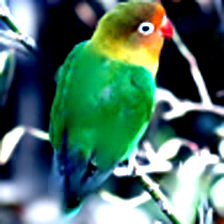}
        \end{subfigure}%
        \caption{Samples from training set}
    \end{subfigure}%
    \hfill
    \begin{subfigure}[t]{0.31\textwidth}
        \centering
        \begin{subfigure}[t]{0.32\textwidth}
        \includegraphics[width=\linewidth]{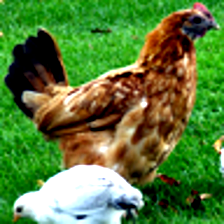}
        \includegraphics[width=\linewidth]{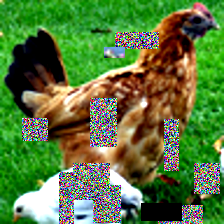}
        \includegraphics[width=\linewidth]{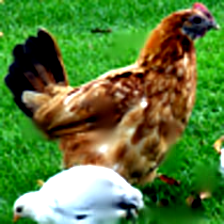}
        \end{subfigure}%
        \hfill
        \begin{subfigure}[t]{0.32\textwidth}
        \includegraphics[width=\linewidth]{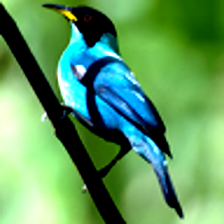}
        \includegraphics[width=\linewidth]{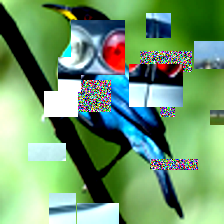}
        \includegraphics[width=\linewidth]{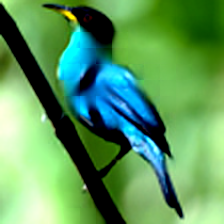}
        \end{subfigure}%
        \hfill
        \begin{subfigure}[t]{0.32\textwidth}
        \includegraphics[width=\linewidth]{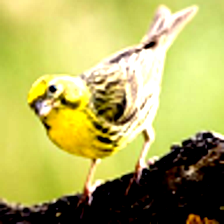}
        \includegraphics[width=\linewidth]{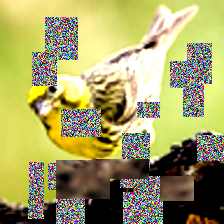}
        \includegraphics[width=\linewidth]{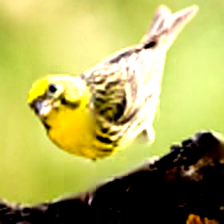}
        \end{subfigure}%
        \caption{Samples from testing set}
    \end{subfigure}%
    \hfill
    \begin{subfigure}[t]{0.32\textwidth}
        \centering
        \begin{subfigure}[t]{0.32\textwidth}
        \includegraphics[width=\linewidth]{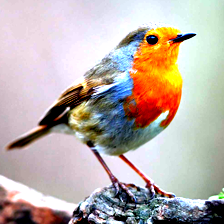}
        \includegraphics[width=\linewidth]{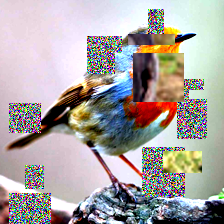}
        \includegraphics[width=\linewidth]{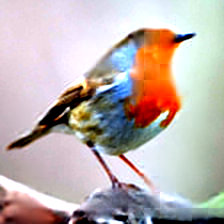}
        \end{subfigure}%
        \hfill
        \begin{subfigure}[t]{0.32\textwidth}
        \includegraphics[width=\linewidth]{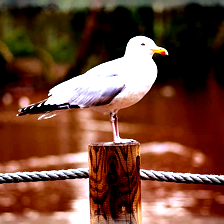}
        \includegraphics[width=\linewidth]{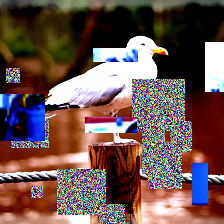}
        \includegraphics[width=\linewidth]{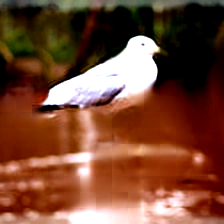}
        \end{subfigure}%
        \hfill
        \begin{subfigure}[t]{0.32\textwidth}
        \includegraphics[width=\linewidth]{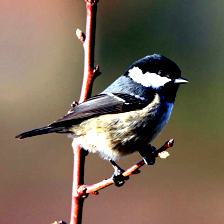}
        \includegraphics[width=\linewidth]{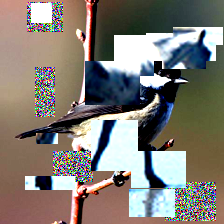}
        \includegraphics[width=\linewidth]{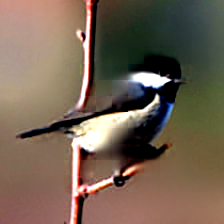}
        \end{subfigure}%
        \caption{Images from internet}
    \end{subfigure}%
    \caption{Reconstructed images from our trained SiT model. The images are randomly obtained from (a) Training data, (b) Test data, and (c) From the internet. Each row refers to the original images, corrupted images, and the reconstructed images, respectively. We applied moderate corruption for demonstration purposes.}
    \label{fig:reconstructionSamples}
\end{figure*}

\subsection{End-to-End Self-Supervised Training}

For a given mini-batch consisting of $N$ samples, two different random augmentations are applied to each sample. Augmented images created from the same samples are considered to constitute positive pairs and images coming from different samples are considered as negative pairs. Next, GMML image-based manipulation is applied to both images. The network is then trained to reconstruct the original image after the applied corruptions and maximise the cosine similarity between the positive pairs within the batch. The overall loss function is shown in Equation \ref{eq:Loss}.

\begin{align}
\begin{split}
\mathcal{L}_{\rm total}(\mathbf{W}) & = \alpha \times\mathcal{L}_{\rm recons}(\mathbf{W}) + \mathcal{L}_{\rm contr}(\mathbf{W})
\end{split}
\label{eq:Loss}
\end{align}

Note that $\alpha$ is the scaling factors of our multi-task objective function. We set $\alpha$ to a high value, i.e. $\alpha=5$ when the model is pretrained on small-scale datasets to enables ViTs to learn the useful local inductive bias. In the case of large-scale datasets, e.g. ImageNet-1K, we set $\alpha=1$.

After the self-supervised training, we apply transfer learning by replacing the contrastive head to output $c$ nodes corresponding to the number of classes in the downstream task.

\section{Experimental Results}
\label{sec:exp}

The common evaluation to demonstrate the generalisation of the learnt features by self-supervised methods is to pretrain the model in unsupervised fashion, followed by finetuning the model on a downstream task like image  classification,  object  detection, segmentation, etc. In this work, we conduct several experiments on different well-known multi-class and multi-label datasets (Table \ref{tbl:dataset}) as well as on video instance segmentation task to show the effectiveness of our proposed self-supervised vision transformer.

\input{tables/Datasets}
\input{tables/Multi_Class_Small}
\input{tables/Multi_Class_INet}

\subsection{Implementation  Details}
We implement the self-supervised architecture using vision transformer \cite{dosovitskiy2020image}, mostly employing the small variant of vision transformers (ViT-S) with $224\times224$ input image size, $16\times16$ patch size, $384$ hidden dimension, $12$ consecutive MSA and MLP blocks, and $6$ heads on each multi-head self-attention layer. There are $21$ million  parameters in total in this architecture.

For the image corruption, we either replace random patches from the image with noise or with patches from other images. The width and height of the corrupted patches varies from 5\% to 25\% of the input image size with the overall replacement rate of up to 70\% in the case of noise replacement and up to 30\% in the case of replacement from other images. The image reconstruction head  consists of 2 fully connected layers with $2048$ neurons and GeLU~\cite{hendrycks2016gaussian} non-linearity each, followed by a transposed convolution to return back to the image space. 

For the contrastive learning, we employed SimCLR \cite{chen2020simple} with the temperature parameter set to $0.2$. The contrastive learning head consists of two fully connected layers with $4096$ neurons, batch normalisation layer, and GeLU non-linearity each, followed by a linear layer with $256$ output nodes that represent the image embeddings. The momentum encoder is updated using exponential moving average of the encoder weights with $\lambda$ following a cosine schedule from $0.996$ to $1$ during training.

For the optimisation of self-supervised models, we trained all the models using the Adam optimiser~\cite{Loshchilov2017FixingWD} with batch size $64$, momentum $0.9$, weight decay $0.05$ and learning rate of $5e^{-4}$ for $800$ epochs in total for ImageNet-1K pretraining and $3000$ epochs for small datasets pretraining. In fact, we mostly rely on the vision transformer developer's default hyper-parameters. We believe that further improvements can be obtained by tuning the hyper-parameters for the self-supervised  model.

Simple data augmentation techniques are applied during the self-supervised training. We found that to learn low-level features as well as higher-level semantic information, aggressive data augmentation like MixUp \cite{zhang2017mixup} and Auto-Augment \cite{cubuk2019autoaugment} hurts the training with the objective functions in hand. Therefore, we used only cropping, horizontal flipping, colour jittering, solarization, and Gaussian blurring. 

For every instance created by augmentation, GMML-based image manipulation described in Section \ref{sec:method} is applied to perturb the image and the network is optimised together with the contrastive learning to reconstruct the image after distortion. In order to  get a feel for the reconstruction capability of SiT, in Figure \ref{fig:reconstructionSamples} we show the reconstruction of randomly selected images from the training data, testing data, and samples from internet after applying moderate corruption to them for demonstration purposes. 

Finally, for the finetuning step, the reconstruction and contrastive heads are dropped and an output layer with $c$ nodes corresponding to the number of classes in the downstream task is appended to the class token of the teacher network. 

\input{tables/Multi-Label}

\subsection{Multi-class Classification}

\subsubsection{Multi-class Classification on Small Datasets}
\label{exp:small_datasets}

It is well known that transformers are data-hungry which make them hard to train, mostly, due to the lack of the typical convolutional inductive bias.
Consequently, the common protocol for self-supervised learning with transformers is to pretrain the model on a large scale dataset, such as ImageNet-1K or even larger datasets. The compute and data demand of the vision transformers limit their adoption, particularly by AI researchers with smaller resource budget. Therefore, in the first set of experiments we investigate the applicability of training transformers from scratch with limited data. Particularly, we compare our proposed SiT approach with the state-of-the-art SSL methods when the pretraining and finetuning are performed only on the target dataset. Unfortunately, few works in the literature show the impact of their proposed methods when pretrained with limited data. To compare with the concurrent and post arts in the literature, we pretrained and finetuned the self-supervised state-of-the-art methods using the publicly available codes and the default parameters as suggested by the authors. 
Particularly, we compare with MoCo-V3 \cite{chen2021empirical}, Dino \cite{caron2021emerging}, MAE \cite{Kaiming2021mae}, and SimMIM \cite{xie2022simmim}. For the finetuning step, we rely on the vision transformer developer's default hyper-parameters ~\cite{touvron2020training}. 

As shown in Table \ref{tab:MC_perf_smalldatasets}, the self-supervised pretraining of SiT consistently enhances the performance on all datasets compared to when trained from scratch with a large margin (up to 64.7\% in the case of Cars dataset).

Further, our method outperforms the concurrent and post arts in SSL with a large margin with an improvement of +3.9\%, +11.7\%, +11.8\%, +8.8\%, +1.1\%, +1.1\%, +0.8\%, and +1.8\% on Flowers, Pets, CUB, Aircraft, STL10, Cars, CIFAR10, and CIFAR100 datasets, respectively. Note that the performance of SimMIM on small datasets is not included in Table \ref{tab:MC_perf_smalldatasets} as the pretrained models did not converge, resulting in poor finetuning performance. Presumably, different recipes might be required to pretrain SimMIM on small datasets. 

In general, contrastive SSL approaches require special data-specific design or hyper-parameter tuning \cite{cao2022training}, making them not suitable for small datasets which justify the poor performance of MoCo-V3 and Dino in Table~\ref{tab:MC_perf_smalldatasets}. The high performance of SiT on small stand-alone datasts is attributed to modelling of local statistical correlations (which is missing in transformers) and global information dictated by the data itself. Even though MAE is also using masked image modelling, the design choice of MAE models the inductive bias mainly in the heavy decoder which is then passed down to last layers of encoder. This counter intuitive way to modelling information makes MAE not suitable for small datasets.

\subsubsection{Multi-class Classification on Large-scale Datasets}
\label{exp:large_datasets}

In this section, we show the effectiveness of SiT when pretrained on large-scale dataset, i.e. ImageNet-1K, and finetuned on several multi-class classification datasets. In Table \ref{tbl:large_scale_classification}, we show that our proposed method outperforms supervised pretraining in most of the datasets, and achieve an improvement of 1.1\% when finetuned on ImageNet-1K dataset (in the case of ViT-S/16). Further, the performance of our proposed method is outperforming or on par with the concurrent and post art methods in most of the small datasets as well as large-scale dataset when it is pretrained employing bigger transformer architecture, i.e. ViT-B/16.

\subsection{Multi-Label Classification}
\label{sec:multilabel}

In Table \ref{tbl:res_multilabel}, we compare the proposed SiT with DeiT \cite{touvron2020training}, MoCo-v3 \cite{chen2021empirical}, and DINO \cite{caron2021emerging} frameworks on three different multi-label datasets, PASCAL VOC, MS-COCO, and Visual-Genome, respectively. 

All the models are finetuned using 1 Nvidia Tesla V100 32GB GPU card for $60$ epochs with $16$ batch size and $480\times480$ input size, employing Adam optimiser and binary cross entropy loss function. For evaluation, we employ the mean average precision (mAP). 

First, we show the results when ViT-S/16 is trained from scratch on the downstream task. Then, we show the performance when the model is pretrained and finetuned employing the same downstream dataset. Finally, we report the accuracy when the models are pretrained with ImageNet-1K employing SiT, MoCo v3, and DINO frameworks, and then finetuned on the multi-label downstream datasets.

From the reported results, it is evident that the training from random initialisation has produced low accuracies as the amount of data available is insufficient to train the transformer. The results significantly improved when the models are pretrained using SiT without any external data with $+40.6$, $+27.1$, and $+8.0$ absolute mAP improvement in PASCAL VOC, MS-COCO, and Visual-Genome datasets, respectively. Further, pretraining with the SiT framework on ImageNet-1K outperforms SSL SOTA frameworks with $+0.5$ and $+0.6$ absolute mAP improvement in MS-COCO and Visual-Genome datasets. 

\subsection{Instance Segmentation}
\label{exp:Ins_seg}

In this experiment, we demonstrate the generalisation of the frozen learnt features by SiT, pretrained on ImageNet-1K, on the DAVIS-2017 video instance segmentation benchmark \cite{pont20172017}. Specifically, the features of the data tokens of the DAVIS-2017 videos are extracted by passing the videos to the frozen pretrained SiT model and then follow the experimental protocol in \cite{jabri2020space} by segmenting the scenes with nearest neighbour between consecutive frames. 

As shown in Table \ref{tbl:segm}, SiT outperforms state-of-the-art supervised and self-supervised methods with a large margin. Note that the poor performance of MAE is expected as MAE representations are less linearly separable than contrastive learning based methods.

\input{tables/Segmentation}

\begin{figure*}[t]
\centering
\begin{minipage}{0.33\textwidth}
\centering
\includegraphics[width=\linewidth]{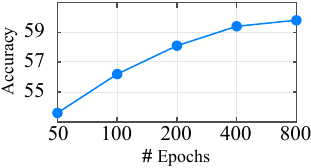}
\caption*{(a)}
\end{minipage}\hfill
\begin{minipage}{0.33\textwidth}
\centering
\includegraphics[width=\linewidth]{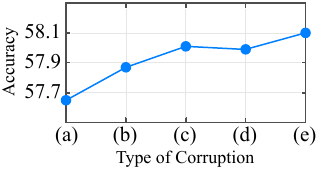}
\caption*{(b)}
\end{minipage}\hfill
\begin{minipage}{0.34\textwidth}
\centering
\includegraphics[width=\linewidth]{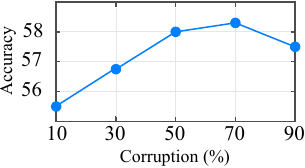}
\caption*{(c)}
\end{minipage}
\vspace{-0.2cm}
\caption{Ablation studies of the effect of (a) Longer pretraining, (b) Type of corruption, and (c) Percentage of corruption.}
\label{fig:ablations}
\end{figure*}

\subsection{Ablation Study}

For the ablation studies, we used the image recognition task defined on a subset of ImageNet-1K dataset as a vehicle for measuring the impact of several configurations of the pretext self-supervised pretraining. Due to the limited resources, all the models are conducted on 10\% of ImageNet-1K for pretraining and evaluated on the full validation set of ImageNet-1K. The models are pretrained for 200 epochs employing the ViT-S/16 architecture as the backbone of the model and the momentum encoder.

\noindent
\textbf{Effect of Different Components of SiT.}
The aim of this ablation study is to investigate the effect of the individual elements of the pretext learning. From the results reported in Table \ref{tbl:SiT_component}, it is evident that the training from random initialisation has produced poor performance of less than 40\%, as the amount of data available is insufficient to train the transformer. We then pretrained SiT using just the contrastive learning head without momentum encoder, which is essentially equivalent to SimCLR \cite{chen2020simple}. The accuracy jumped to 56.8 with an improvement of 19.5\%. Subsequently, we included a momentum encoder (similar to MoCo-v3 \cite{chen2021empirical}) which further improved the accuracy with around 1\%. 

After that, we investigated the effectiveness of training transformers as an autoencoder to reconstruct the input image, i.e. D(E(x)) = x, where x is the input image, E is the encoder which is ViT-S/16 in our case, and D is a lightweight reconstruction decoder. Expectedly, the performance was similar to the performance of a model trained from scratch. Indeed, this is attributed to the fact that without proper choice of constraints, autoencoders are capable of learning identity mapping, i.e. memorising the input without learning any useful discriminative features.
To regularise the transformer-based autoencoder, we incorporated GMML version of the transformer working with the reconstruction error loss function. The performance jumped to 57.5\% which is even better than using contrastive learning with momentum.

Finally, we performed an experiment of using both contrastive learning and GMML, without momentum encoder. The pairing of the self-supervision training mechanisms had  an egalitarian effect, pushing the finetuning test performance to 57.9\%. This was slightly improved by employing a momentum encoder.

In summary, using the reconstruction loss on its own as a means of self-supervision provided an effective starting point for efficient downstream task finetuning. Further marginal improvements can be made by extending the range of mechanisms for self-supervised pretraining. 

\begin{table}[t]
\caption[]{Effect of the different components of SiT for self-supervised pretraining. Models are pretrained on $10\%$ of ImageNet-1K dataset for 200 epochs, followed by finetuning on validation set of ImageNet-1K.}
\label{tbl:SiT_component}
\resizebox{0.99\linewidth}{!}{
\begin{tabular}{lx{1cm}x{1.3cm}x{1.3cm}x{1cm}|x{1.2cm}}
\hline
Method & 
{Contrastive} Head & Momentum Encoder & Image  Corruption& Recons. Head & Top-1 Accuracy\\ \hline
Random init.     
& \xmark & \xmark & \xmark & \xmark & 37.3\\
SimCLR \cite{chen2020simple}      
& \cmark & \xmark & \xmark & \xmark & 56.2\\
MoCo-v3 \cite{chen2021empirical}  
& \cmark & \cmark & \xmark & \xmark & 57.0\\ \cline{2-6}
\multirow{4}{*}{SiT} 
& \xmark & \xmark & \xmark & \cmark & 39.5\\
& \xmark & \xmark & \cmark & \cmark & 57.5\\
& \cmark & \xmark & \cmark & \cmark & 57.9\\   
& \cmark & \cmark & \cmark & \cmark & 58.1\\\hline  
\end{tabular}
}
\end{table}

\noindent
\textbf{Effect of Longer Self-supervised Pretraining.}
In Figure \ref{fig:ablations}-a, we show the quality of SiT when pretrained for longer number of epochs. We observe that the performance steadily increases with the number of epochs and it does not saturate even after pretraining the model for $800$ epochs.

\noindent
\textbf{Effect of Type of Corruption.}
In Figure \ref{fig:ablations}-b, we show the effect of different type of corruption, i.e. corrupting the images with (a) zeros, (b) noise, (c) replace with patches from another image, (d) either by zeros or replacing from another image, and (e) either by noise or replacing from another image. We found that the best individual inpainting task is ``replace'' where connected patches are randomly replaced with patches from another image. Further, we obtained a better performance when ``replace'' is combined with ``noise''. On the other hand, the accuracy slightly dropped when ``zeros'' combined with ``replace''.

\noindent
\textbf{Effect of the Percentage of Image Corruption.}
Figure \ref{fig:ablations}-c shows the top-1 accuracy when the models are pretrained with different corruption percentages. We found that the optimal ratio is between 50\% to 70\%. This behaviour was expected as the masking encourages the network to learn semantic information from the uncorrupted patches surrounding the groups of masked tokens in order to recover the missing information. 

\noindent
\textbf{Effect of Aligning the Corruption with the Patch size.}
We observed that the speed of convergence and generalisation of pretraining slightly improve when the masking of tokens is not aligned to patch boundaries especially when pretraining on small datasets.

\begin{figure*}[t]
    \centering
    \begin{minipage}{0.03\linewidth} \centering \vspace{-4cm} \rotatebox{90}{\textbf{Pretraining}}\end{minipage}
    \begin{subfigure}[t]{0.31\textwidth}
        \centering
        \includegraphics[width=\linewidth]{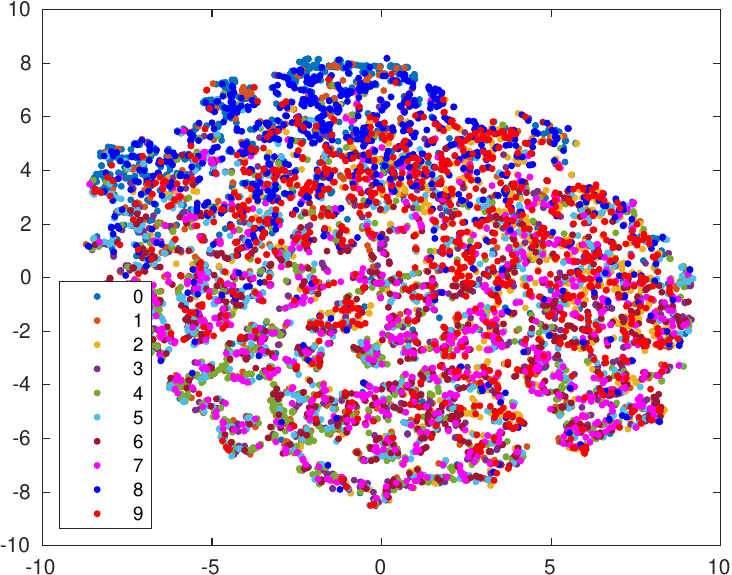}\caption{Random Initialisation}
    \end{subfigure}%
    \hspace{0.01\linewidth}
    \begin{subfigure}[t]{0.31\textwidth}
        \centering
        \includegraphics[width=\linewidth]{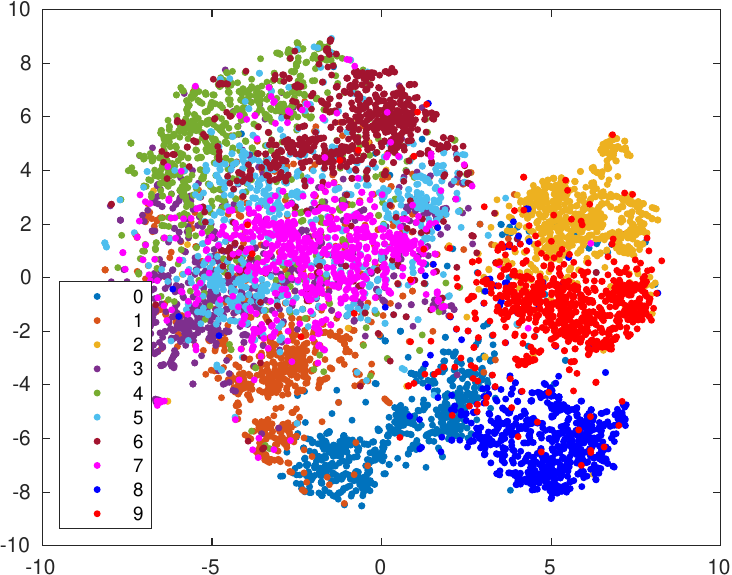}\caption{Pretrained [Labeled data]}
    \end{subfigure}%
    \hspace{0.01\linewidth}
    \begin{subfigure}[t]{0.31\textwidth}
        \centering
        \includegraphics[width=\linewidth]{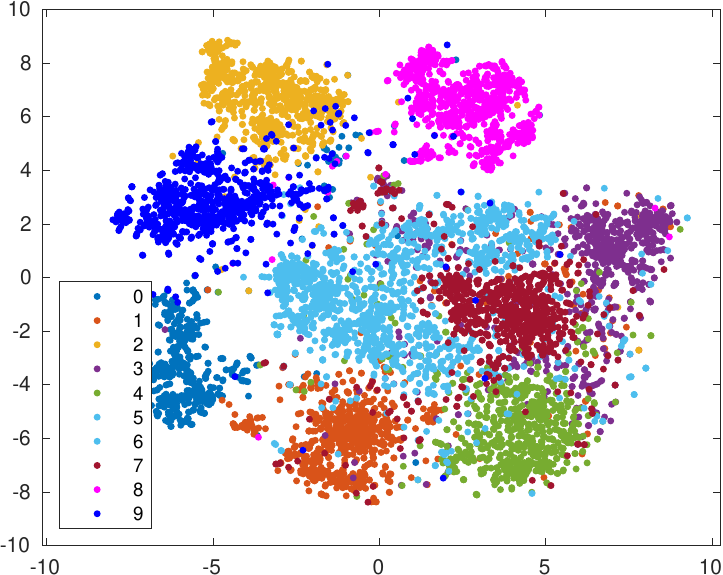}\caption{Pretrained [Labeled+Unlabeled data]}
    \end{subfigure}%

    \begin{minipage}{0.03\linewidth} \centering \vspace{-4cm} \rotatebox{90}{\textbf{Finetuning}}\end{minipage}
    \begin{subfigure}[t]{0.31\textwidth}
        \centering
        \includegraphics[width=\linewidth]{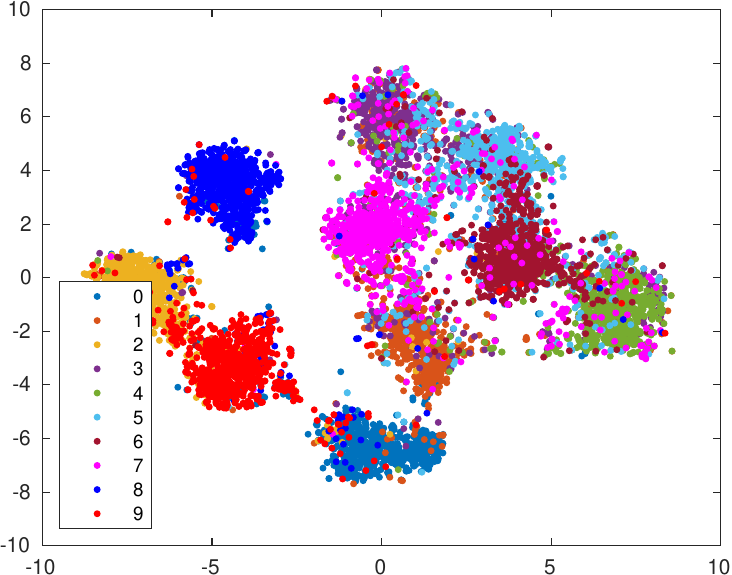}\caption{Training from Scratch}
    \end{subfigure}%
    \hspace{0.01\linewidth}
    \begin{subfigure}[t]{0.31\textwidth}
        \centering
        \includegraphics[width=\linewidth]{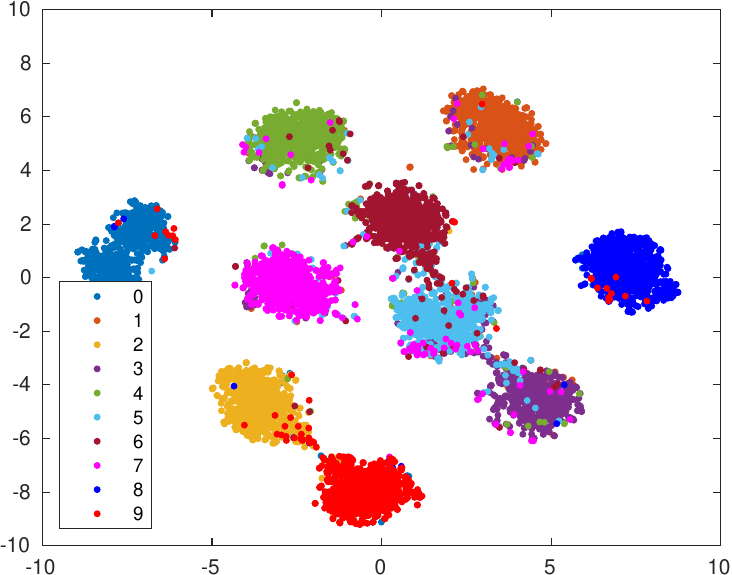}\caption{Finetuning from (b)}
    \end{subfigure}%
    \hspace{0.01\linewidth}
    \begin{subfigure}[t]{0.31\textwidth}
        \centering
        \includegraphics[width=\linewidth]{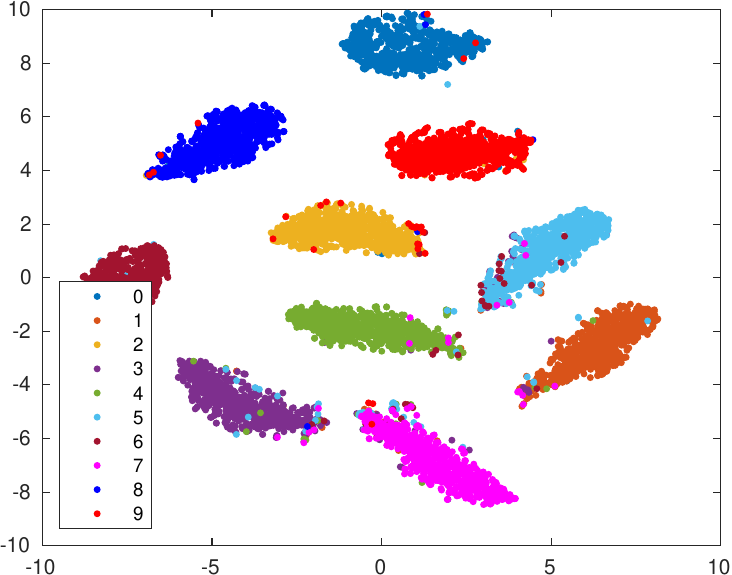}\caption{Finetuning from (c)}
    \end{subfigure}%

    \vspace{-0.2cm}
    \caption{t-SNE visualisation of the embeddings of the 8,000 test set images of STL-10 dataset extracted from the pretrained and finetuned SiT models.}

    \label{fig:representation_visualization}

\end{figure*}

\input{visualisation}

\noindent
\textbf{Qualitative Results.} To gain additional insight and a better understanding of the effect of pretraining the SiT model using unlabeled data, we visualise the t-SNE \cite{van2008visualizing} projection of the learnt representations in different scenarios. In Figure \ref{fig:representation_visualization}-a, we visualise the projection of the embeddings of the test set of STL-10 dataset before training (the model is randomly initialised). Following, in Figure \ref{fig:representation_visualization}-b, we show the projection of the embeddings when SiT is pretrained in unsupervised fashion for 3000 epochs using only the labeled training set of STL-10 dataset that consists of 5,000 images. Notice that even without the label information, SiT is able to discriminate between different classes. Finally, in Figure \ref{fig:representation_visualization}-c, the projection of the embeddings are shown when the model is pretrained using the labeled and unlabeled sets of STL-10 dataset which in total consists of 105,000 images for 1,000 epochs. It is evident that with more unlabeled data in the pretraining stage, the pretrained model has better discrimination ability between the different classes. We also show the t-SNE visualisation after the finetuning stage in Figure \ref{fig:representation_visualization}-d, \ref{fig:representation_visualization}-e, and \ref{fig:representation_visualization}-f. The models are finetuned in the traditional supervised fashion with top-1 validation accuracy of 76\%, 92\%, and 96.5\%, respectively.

\noindent
\textbf{Attention Visualisation.} In Figure \ref{fig:vis}, we provide  visualisations of the attention of the class token after the pretraining stage of SiT employing the small variant of image transformers, i.e. ViT-S/16. The images are randomly selected from the validation set of ImageNet dataset which are not used during the pretraining stage of SiT.

\section{Conclusion}
\label{sec:conc}
In this work we present a self-supervised vision transformer,  trained with unlabelled data to perform pretext tasks, and used the pretrained model as initialisation for finetuning for a downstream classification task. We proposed to use transformers as an autoencoder, which is realisable by using a 2-layer perceptron at the output (thanks to the transformer architecture). We leveraged the attractive property of the transformer architecture of being particularly suited for combining different loss functions along with reconstruction loss. We added an extra token for contrastive learning along with reconstruction loss. The proposed SiT outperformed state-of-the-art self-supervised methods with wide margins. 
This work focused on image classification as a downstream task. We believe that the SiT is admirably suitable for many other downstream tasks like segmentation and detection, however, this conjecture is left for future investigation.

\small{
\bibliographystyle{unsrt}
\bibliography{references}
}

\ifCLASSOPTIONcaptionsoff
  \newpage
\fi


\begin{IEEEbiography}[{\includegraphics[width=1in,height=1.25in,clip,keepaspectratio]{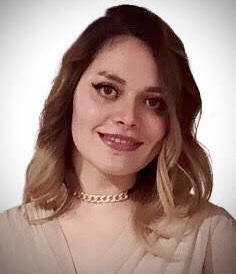}}]{Sara Atito Ali Ahmed} received her Bsc. in computer science from Ain Shams University, Egypt, in 2011. Her Msc. degree was a collaboration between Nile University, Egypt and TU Berlin, Germany in 2014. She received her PhD from Sabanci University (2021), with a thesis on deep learning ensembles  for image understanding. Currently, she is a research fellow in Centre for Vision, Speech and Signal Processing (CVSSP), University of Surrey, UK, working on developing generic self-supervised learning approaches. 
\end{IEEEbiography}
 
\begin{IEEEbiography}[{\includegraphics[width=1in,height=1.25in,clip]{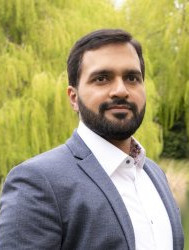}}]{Muhammad Awais}
received the B.Sc. degree in Mathematics and Physics from the AJK University in 2001, B.Sc. degree in computer engineering from UET Taxila in 2005, M.Sc in signal processing and machine intelligence and PhD in machine learning from the University of Surrey in 2008 and 2011.
He is currently a senior lecturer in trustworthy and responsible AI at Surrey Institute for People-Centred Artificial Intelligence and Centre for Vision, Speech and Signal Processing (CVSSP).
His research interests include machine learning, deep learning, self(un,semi)-supervised learning, NLP, audio-visual analysis, medical image analysis and computer vision.
\end{IEEEbiography}

\begin{IEEEbiography}[{\includegraphics[width=1in,height=1.25in,clip,keepaspectratio]{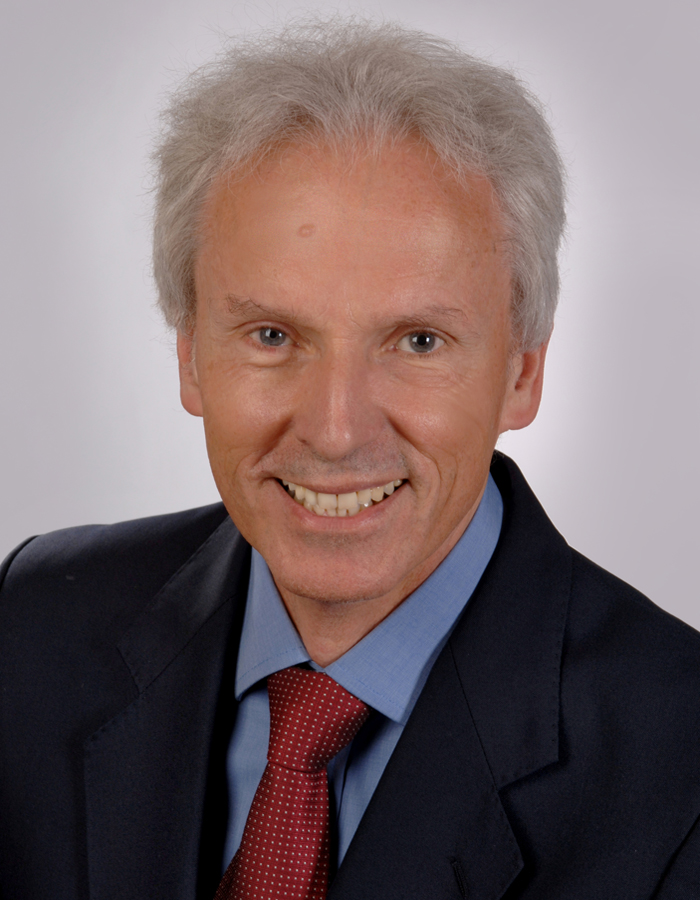}}]{Josef Kittler}
(M'74-LM'12) received the B.A., Ph.D., and D.Sc. degrees from the University of Cambridge, in 1971, 1974, and 1991, respectively.
He is a distinguished Professor of Machine Intelligence at the Centre for Vision, Speech and Signal Processing, University of Surrey, Guildford, U.K.
He conducts research in biometrics, video and image dataset retrieval, medical image analysis, and
cognitive vision. He published the textbook Pattern Recognition: A Statistical Approach and over 700 scientific papers. 
His publications have been cited more than 68,000 times (Google Scholar).

He is series editor of Springer Lecture Notes on Computer Science. He currently serves on the Editorial Boards of Pattern Recognition Letters, Pattern Recognition and Artificial Intelligence, Pattern Analysis and Applications. He also served as a member of the Editorial Board of IEEE Transactions on Pattern Analysis and Machine Intelligence during 1982-1985. He served on the Governing Board of the International Association for Pattern Recognition (IAPR) as one of the two British representatives during the period 1982-2005, President of the IAPR during 1994-1996. Currently he is a member of the KS Fu Prize Committee of IAPR.
\end{IEEEbiography}

\end{document}

%% file: tables/Datasets.tex
\begin{table}[h]
\centering
\caption{Statistics of the employed datasets.}
\label{tbl:dataset}
\begin{tabular}{p{2.5cm} x{1.2cm} x{1.3cm} x{1.2cm}}
\hline
Dataset & \# Classes & 
\# Training Samples & 
\# Testing Samples \\ \hline
&\multicolumn{3}{c}{Multi-class datasets} \\ \cline{2-4}
Flowers  \cite{Flowersdataset}        & 102 & 2,040 & 6,149\\ 
Pets     \cite{PetsDataset}           & 37  & 3,680 & 3,669\\ 
STL10 \cite{coates2011analysis}       & 10  & 5,000 & 8,000\\ 
CUB200 \cite{CUBDataset}              & 200 & 5,994 & 5,794\\ 
Aircraft \cite{AircraftDataset}       & 100 & 6,667 & 3,333\\  
Cars   \cite{Carsdataset}             & 196 & 8,144 & 8,041\\  
CIFAR10 \cite{krizhevsky2009learning} & 10  & 50,000& 10,000\\ 
CIFAR100 \cite{krizhevsky2009learning}& 100 & 50,000& 10,000\\ 
ImageNet-1K\cite{deng2009imagenet}    &1,000&1.28M &50,000\\ \hline
&\multicolumn{3}{c}{Multi-label datasets} \\ \cline{2-4}
Pascal VOC \cite{everingham2015pascal}&20&5,011&4,952\\
MS-COCO \cite{lin2014microsoft} &80&82,081&40,137\\
Visual-Genome \cite{krishna2017visual} &500&98,249&10,000\\ \hline
\end{tabular}
\end{table}

%% file: tables/Multi_Class_Small.tex
\begin{table*}[t]
\centering
\caption{Comparison with state-of-the-art methods when pretrained and finetuned on the target dataset, i.e. no external data is used, employing ViT-S/16.}
\label{tab:MC_perf_smalldatasets}
\resizebox{0.97\linewidth}{!}{
\begin{tabular}{p{2.4cm}x{1.4cm}x{1.2cm}x{1.2cm}x{1.5cm}x{1.2cm}x{1.2cm}x{1.2cm}x{1.3cm}}
\hline
Method & Flowers & Pets & CUB & Aircraft & STL10 & Cars & CIFAR10 & CIFAR100 \\ \hline
\textit{Random init.} & 68.8 & 47.5 & 25.3 & 31.1 & 77.1 & 27.4 & 96.9 & 77.8 \\
\multicolumn{9}{l}{\textit{\color{gray}{{Comparison with concurrent works}}}}\\ 
MoCo-v3 \cite{chen2021empirical}& 88.9 & 69.0 & 53.1 & 62.5 & 95.4 & 84.0 & 97.3 & 83.4\\
\multicolumn{9}{l}{\textit{\color{gray}{{Comparison with post arts}}}}\\ 
Dino \cite{caron2021emerging}   & 82.4 & 58.0 & 43.6 & 49.3 & 92.1 & 73.0 & 96.8 & 78.9\\
MAE \cite{Kaiming2021mae}     &86.9& 73.0 &59.4 & 69.0 & -- & 91.0 & -- & --\\
SiT                           &\textbf{92.8}& \textbf{84.7} & \textbf{71.2} & \textbf{77.8} & \textbf{96.5} & \textbf{92.1} & \textbf{98.2} & \textbf{85.2}\\ \hline
\end{tabular}
}
\end{table*}

%% file: tables/Multi_Class_INet.tex
\begin{table*}[t]
\centering
\caption{Domain Transfer of SiT pretrained on ImageNet-1K dataset.}
\label{tbl:large_scale_classification}
\resizebox{\linewidth}{!}{
\begin{tabular}{p{2.6cm} x{1.4cm}x{1.0cm}x{1.0cm}x{1.0cm}x{1.2cm}x{1.0cm}x{1.2cm}x{1.3cm} | x{1.9cm}}
\hline
Method & Flowers & Pets & CUB & Aircraft & STL10 & Cars & CIFAR10 & CIFAR100 & ImageNet-1K \\ \cline{2-10}
& \multicolumn{9}{c}{ViT-S/16} \\ \cline{2-10} 
\textit{Random init.} & 
68.8 & 47.5 & 25.3 & 31.1 & 77.1 & 27.4 & 96.9 & 77.8 & --\\
Supervised \cite{touvron2020training} &
98.1 & 91.1 & 82.7 & 80.8 & 98.2 & 91.7 & 98.3 & 86.9 & 79.9\\
\multicolumn{9}{l|}{\textit{\color{gray}{{Comparison with concurrent works}}}}\\ 
MoCo-v3 \cite{chen2021empirical} & 
97.7 & 92.3 & 82.6 & 87.3 & 98.0 & 93.0 & 98.2 & 86.6 & 81.4\\
Dino* \cite{caron2021emerging} &
97.8 & 89.4 & 80.8 & 83.8 & 96.7 & 93.1 & 98.6 & 87.1 & 81.5\\
SiT &
\textbf{98.2} & \textbf{92.6} & \textbf{84.6} & \textbf{87.6} & \textbf{98.8} & \textbf{93.2} & \textbf{99.0} & \textbf{90.8} & \textbf{82.0}\\ \hline
& \multicolumn{9}{c}{ViT-B/16} \\ \cline{2-10} 
\multicolumn{9}{l|}{\textit{\color{gray}{{Comparison with concurrent works}}}}\\ 
MoCo-v3 \cite{chen2021empirical} &
98.3 & \textbf{93.7} & 84.1 & 87.2 & \textbf{98.4} & 93.4 & 98.2 & 87.3 & 83.2\\
Dino \cite{caron2021emerging}&
98.4&90.2&80.7&81.5& 97.2 &93.0&98.2 &87.1 &82.8\\
\multicolumn{9}{l|}{\textit{\color{gray}{{Comparison with post arts}}}}\\
SimMIM* \cite{xie2022simmim} &
97.2&92.3&81.8&83.4&97.8&91.5&98.9&88.6&\textbf{\underline{83.6}}\\
MAE* \cite{Kaiming2021mae}    &
\textbf{98.9} &92.8 &84.2 &\textbf{88.4} & 98.2 &93.5& \textbf{99.0} & -- &\textbf{\underline{83.4}}\\
SiT & 98.4 & 93.0 & \textbf{84.6} & 88.0 & \textbf{98.4} & \textbf{93.7} & \textbf{99.0} & \textbf{89.2} & \textbf{\underline{83.5}}\\ \hline
\end{tabular}
}
\end{table*}

%% file: tables/Multi-Label.tex
\begin{table*}[!t]
\centering
\caption{mAP (mean Average Precision) of regular inference on the PASCAL VOC
2007, VG-500, and MS-COCO datasets. * are run by us using official pre-trained weights. All the models are pre-trained using ViT-S/16 vision transformer (unless mentioned otherwise) with $224\times224$ input resolutions and supervised finetuning with $448\times448$ resolution.}
\label{tbl:res_multilabel}
\resizebox{\linewidth}{!}{
\begin{tabular}{p{8cm}x{2cm}x{2cm}x{2cm}}
\hline
Method  & PASCALVOC & MSCOCO & Visual-Genome\\ \hline
& \multicolumn{3}{c}{Pre-training using the given dataset}\\ \cline{2-4}
Supervised training from scratch &  34.1 & 47.9 & 23.7\\ 
SSL pre-training (SiT) & \textbf{74.7} & \textbf{75.0} & \textbf{31.7}\\\hline
& \multicolumn{3}{c}{Pre-training using ImageNet-1K dataset}\\ \cline{2-4}
Supervised pre-training (ResNet-101 \cite{he2016deep})        & \textbf{92.9} & 78.6 & 30.9\\
Supervised pre-training (ViT-S/16 \cite{touvron2020training})*& 92.6 & 81.4 & 33.0\\ \cline{2-4}
SSL pre-training ($\text{DeepCluster}-\text{\small{[AlexNet]}}$ \cite{caron2018deep}) & 73.7 & -- & --\\
SSL pre-training ($\text{SimCLR}-\text{\small{[ResNet-50]}}$ \cite{chen2020simple})   & 84.1 & -- & --\\
SSL pre-training (MoCo v3 \cite{chen2021empirical})*      & 86.0 & 77.8 & 32.3\\
SSL pre-training ($\text{Dino}$ \cite{caron2021emerging})* & 91.6 & 80.8 & 33.4\\
SSL pre-training ($\text{SiT}$)                   & 92.0 & \textbf{81.9} & \textbf{34.0}\\ \hline
\end{tabular}
}
\end{table*}

%% file: tables/Segmentation.tex
\begin{table}[h!]
\centering
\caption{DAVIS 2017 Video object segmentation. We report the mean region similarity ($J_m$) and the mean contour-based accuracy ($F_m$) on video instance tracking. All the models are pretrained using ViT-S/16 vision transformer (unless mentioned otherwise) with $224\times224$ input resolutions on ImageNet-1K dataset and evaluated with $448\times448$ resolution on DAVIS-2017 dataset.}
\label{tbl:segm}
\begin{tabular}{lccccc}
Method     &  Architecture & Data &$(J\&F)_m$    & $J_m$    & $F_m$   \\\hline
\multicolumn{6}{l}{\textit{\color{gray}{{Supervised Learning}}}}\\
Deit \cite{touvron2020training} &  ViT-S/16 & INet & 57.3   & 55.6  &   59.0     \\\hline
\multicolumn{6}{l}{\textit{\color{gray}{{Self-Supervised Learning}}}}\\
MoCo-v3 \cite{chen2021empirical} & \multirow{3}{*}{ViT-S/16} & \multirow{3}{*}{INet} &
           56.0 & 53.4 & 58.5 \\
DINO \cite{caron2021emerging}    &&&61.8 & 60.2 & 63.4   \\
SiT     &&&\textbf{63.1} & \textbf{61.3} & \textbf{64.9}   \\\hline
\color{gray}{MAE \cite{Kaiming2021mae}} &\color{gray}{ViT-B/16} & \color{gray}{INet} & \color{gray}{51.0} & \color{gray}{49.4} & \color{gray}{52.6}  \\
\end{tabular}
\end{table}

%% file: visualisation.tex
\begin{figure*}[t!]
    \centering
    
    \begin{subfigure}[t]{\linewidth}
    \centering
    
        \begin{subfigure}[t]{0.085\linewidth}
        \includegraphics[width=\textwidth]{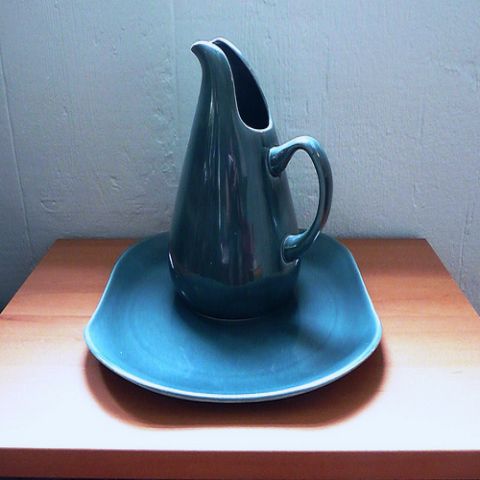}
        \end{subfigure}
        \begin{subfigure}[t]{0.085\linewidth}
        \includegraphics[width=\textwidth]{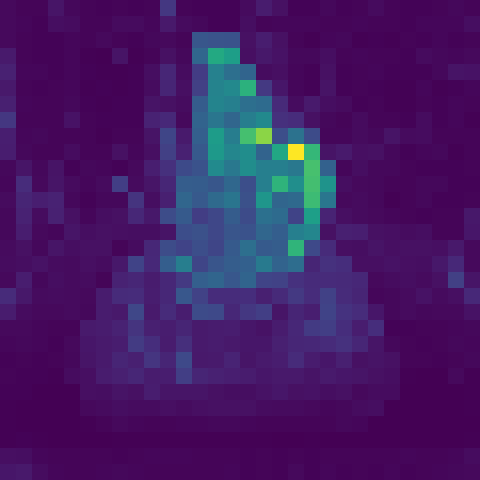}
        \end{subfigure}
        \hspace{0.15cm}
        \begin{subfigure}[t]{0.085\linewidth}
        \includegraphics[width=\textwidth]{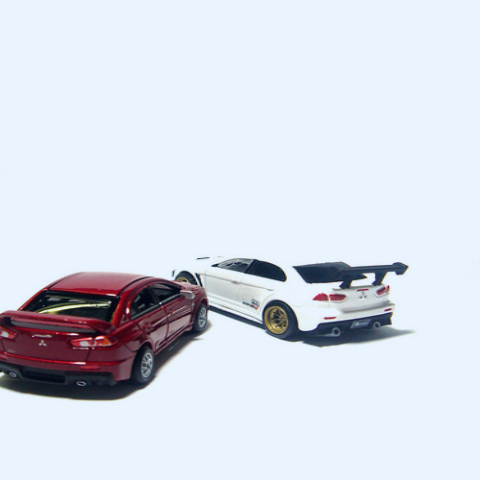}
        \end{subfigure}
        \begin{subfigure}[t]{0.085\linewidth}
        \includegraphics[width=\textwidth]{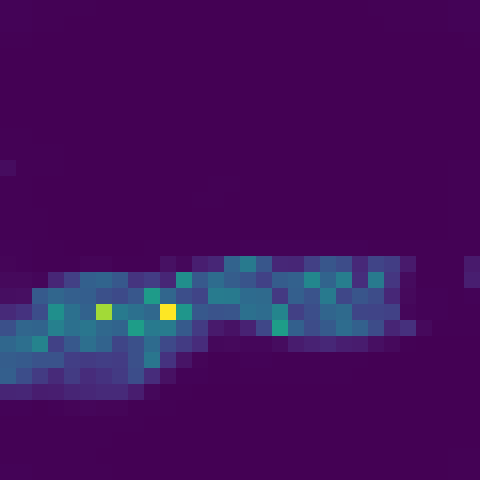}
        \end{subfigure}
        \hspace{0.15cm}
        \begin{subfigure}[t]{0.085\linewidth}
        \includegraphics[width=\textwidth]{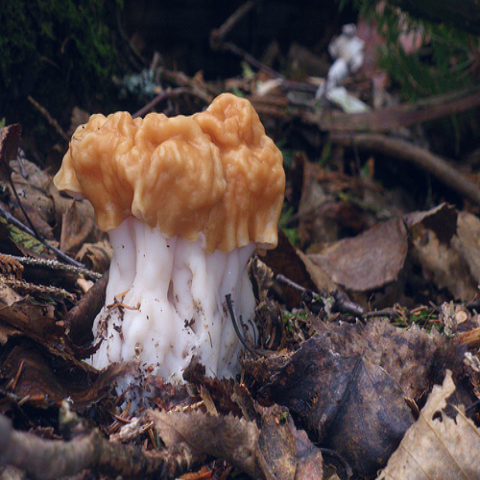}
        \end{subfigure}
        \begin{subfigure}[t]{0.085\linewidth}
        \includegraphics[width=\textwidth]{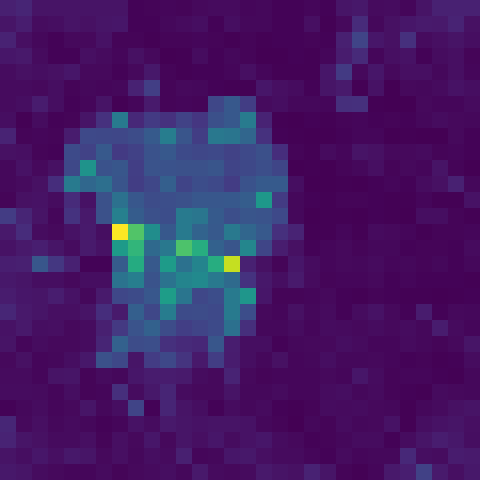}
        \end{subfigure}
        \hspace{0.15cm}
        \begin{subfigure}[t]{0.085\linewidth}
        \includegraphics[width=\textwidth]{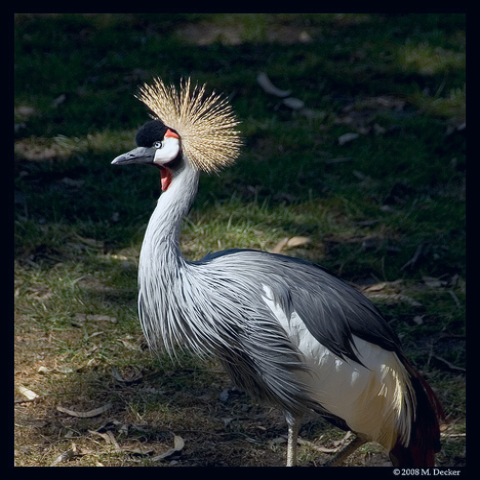}
        \end{subfigure}
        \begin{subfigure}[t]{0.085\linewidth}
        \includegraphics[width=\textwidth]{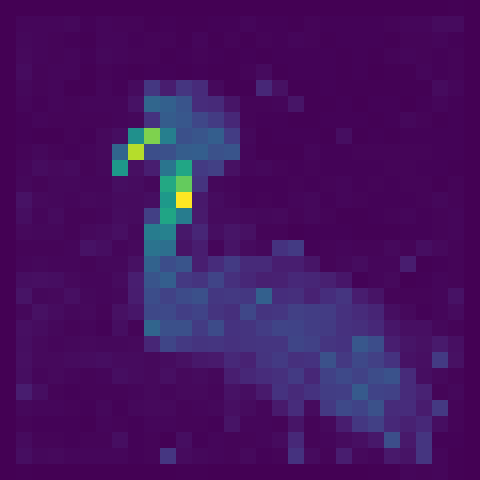}
        \end{subfigure}
        \hspace{0.15cm}
        \begin{subfigure}[t]{0.085\linewidth}
        \includegraphics[width=\textwidth]{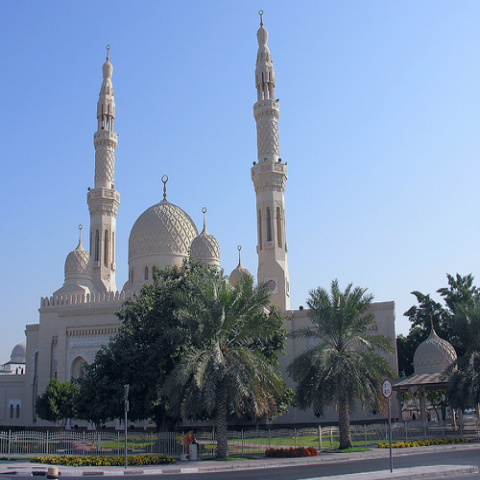}
        \end{subfigure}
        \begin{subfigure}[t]{0.085\linewidth}
        \includegraphics[width=\textwidth]{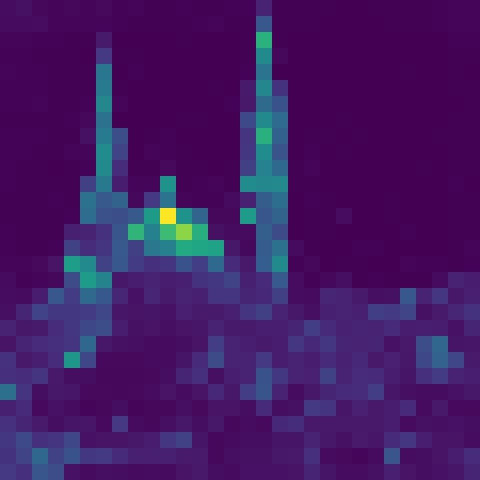}
        \end{subfigure}
        
    \end{subfigure}

    \begin{subfigure}[t]{\linewidth}
    \centering
    
        \begin{subfigure}[t]{0.085\linewidth}
        \includegraphics[width=\textwidth]{images/visualisation/ILSVRC2012_val_00002002.png}
        \end{subfigure}
        \begin{subfigure}[t]{0.085\linewidth}
        \includegraphics[width=\textwidth]{images/visualisation/ILSVRC2012_val_00002002_attn-sum.png}
        \end{subfigure}
        \hspace{0.15cm}
        \begin{subfigure}[t]{0.085\linewidth}
        \includegraphics[width=\textwidth]{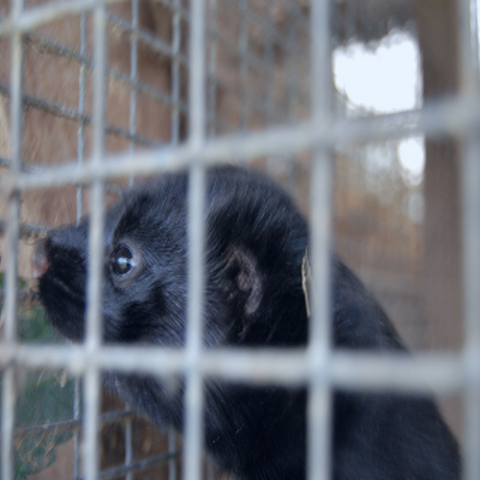}
        \end{subfigure}
        \begin{subfigure}[t]{0.085\linewidth}
        \includegraphics[width=\textwidth]{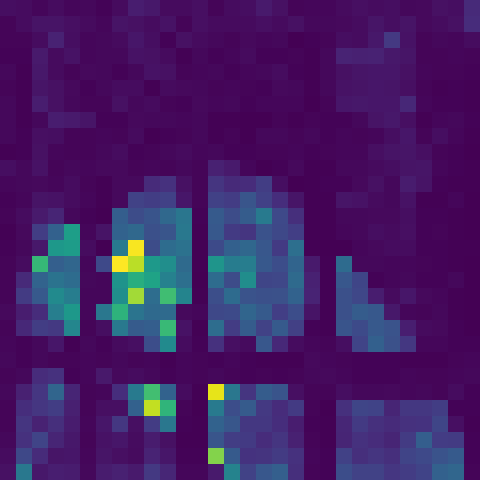}
        \end{subfigure}
        \hspace{0.15cm}
        \begin{subfigure}[t]{0.085\linewidth}
        \includegraphics[width=\textwidth]{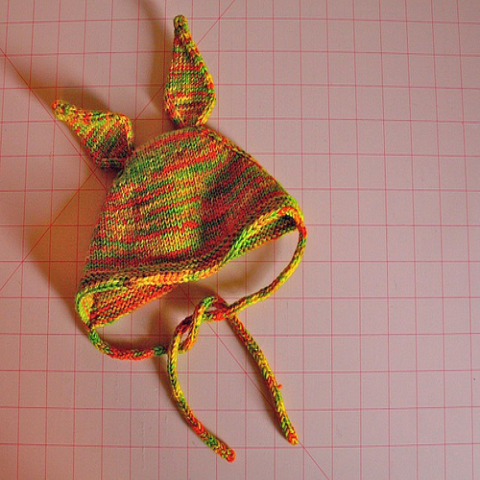}
        \end{subfigure}
        \begin{subfigure}[t]{0.085\linewidth}
        \includegraphics[width=\textwidth]{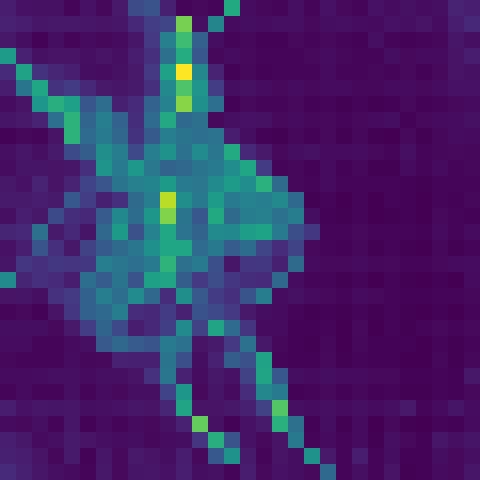}
        \end{subfigure}
        \hspace{0.15cm}
        \begin{subfigure}[t]{0.085\linewidth}
        \includegraphics[width=\textwidth]{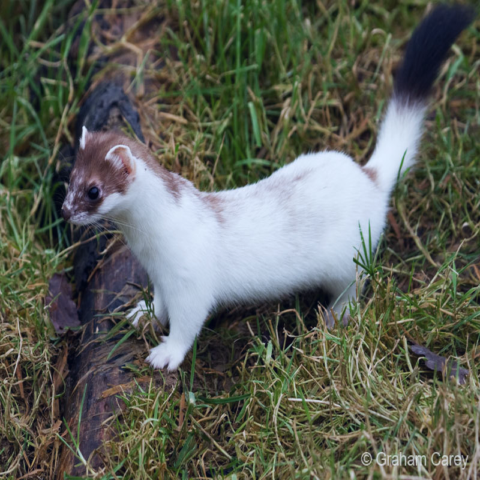}
        \end{subfigure}
        \begin{subfigure}[t]{0.085\linewidth}
        \includegraphics[width=\textwidth]{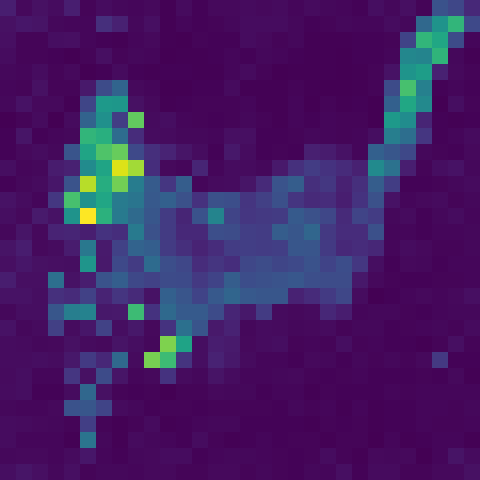}
        \end{subfigure}
        \hspace{0.15cm}
        \begin{subfigure}[t]{0.085\linewidth}
        \includegraphics[width=\textwidth]{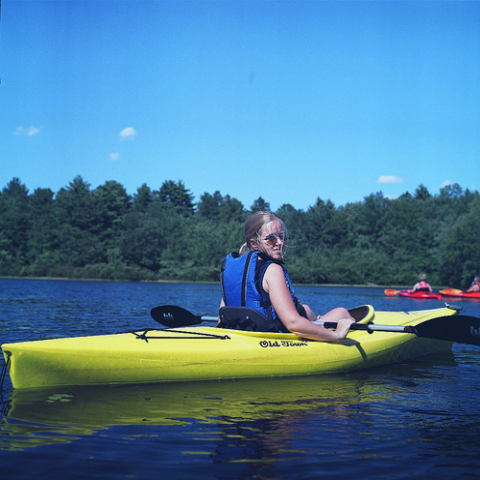}
        \end{subfigure}
        \begin{subfigure}[t]{0.085\linewidth}
        \includegraphics[width=\textwidth]{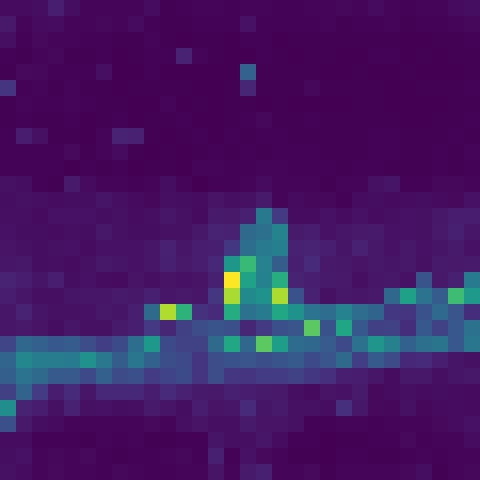}
        \end{subfigure}
        
    \end{subfigure}
    
    \begin{subfigure}[t]{\linewidth}
    \centering
    
        \begin{subfigure}[t]{0.085\linewidth}
        \includegraphics[width=\textwidth]{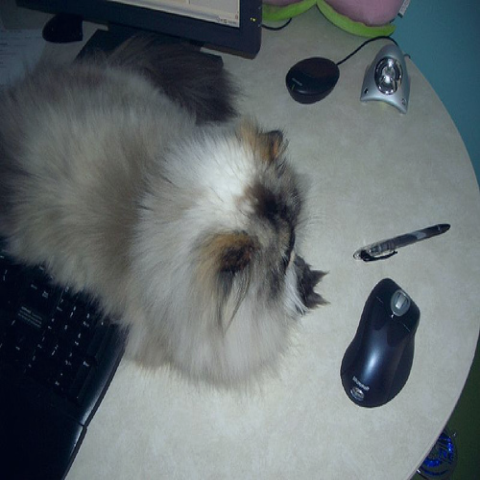}
        \end{subfigure}
        \begin{subfigure}[t]{0.085\linewidth}
        \includegraphics[width=\textwidth]{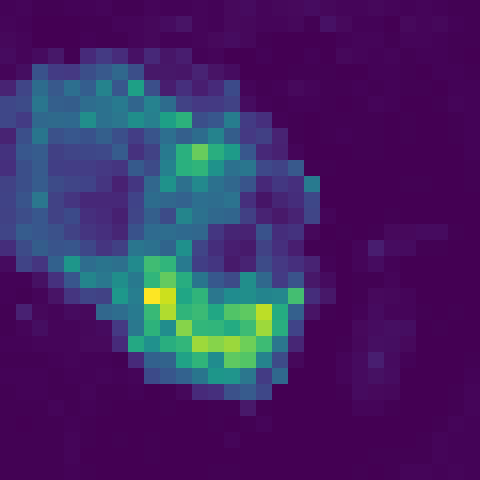}
        \end{subfigure}
        \hspace{0.15cm}
        \begin{subfigure}[t]{0.085\linewidth}
        \includegraphics[width=\textwidth]{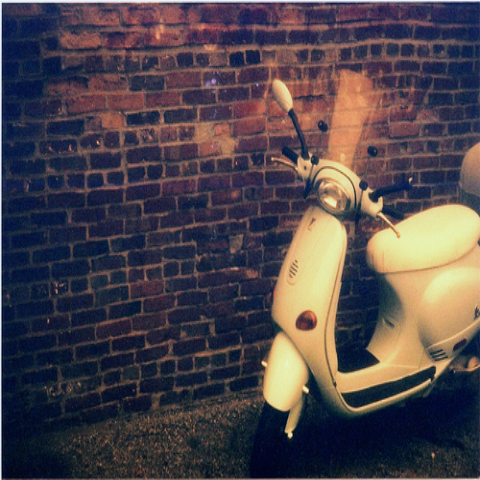}
        \end{subfigure}
        \begin{subfigure}[t]{0.085\linewidth}
        \includegraphics[width=\textwidth]{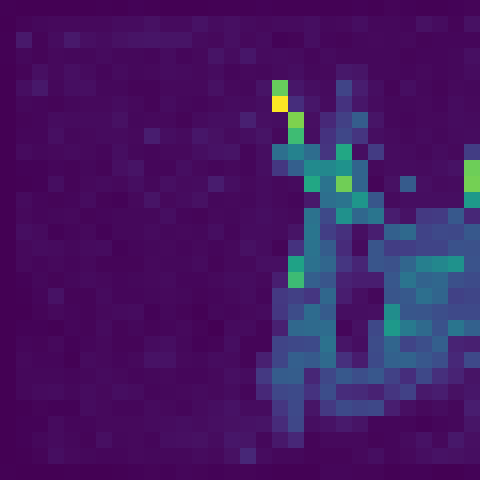}
        \end{subfigure}
        \hspace{0.15cm}
        \begin{subfigure}[t]{0.085\linewidth}
        \includegraphics[width=\textwidth]{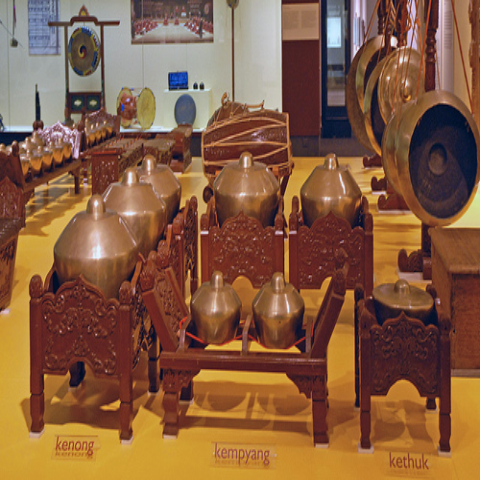}
        \end{subfigure}
        \begin{subfigure}[t]{0.085\linewidth}
        \includegraphics[width=\textwidth]{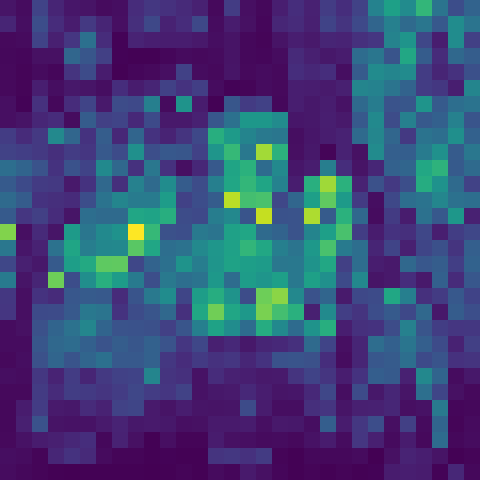}
        \end{subfigure}
        \hspace{0.15cm}
        \begin{subfigure}[t]{0.085\linewidth}
        \includegraphics[width=\textwidth]{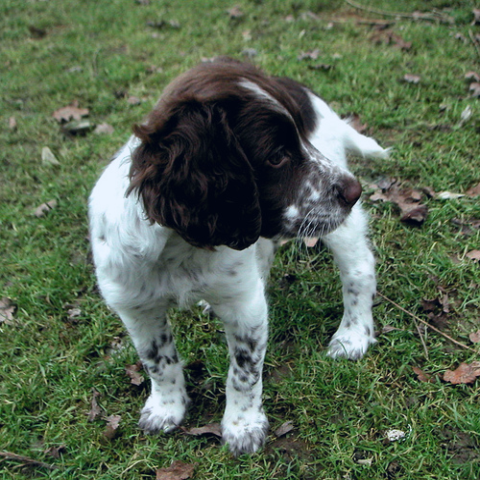}
        \end{subfigure}
        \begin{subfigure}[t]{0.085\linewidth}
        \includegraphics[width=\textwidth]{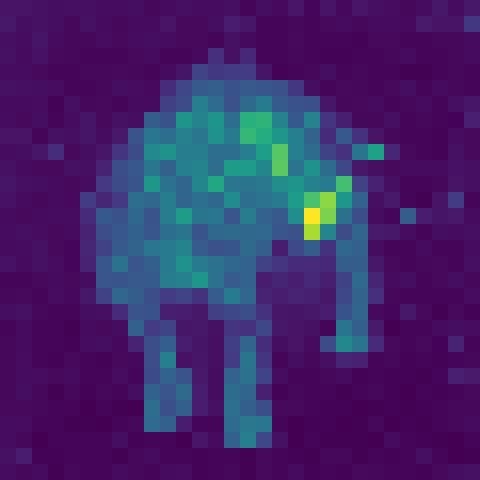}
        \end{subfigure}
        \hspace{0.15cm}
        \begin{subfigure}[t]{0.085\linewidth}
        \includegraphics[width=\textwidth]{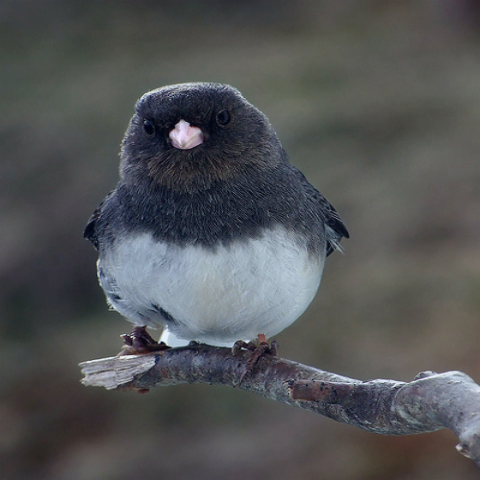}
        \end{subfigure}
        \begin{subfigure}[t]{0.085\linewidth}
        \includegraphics[width=\textwidth]{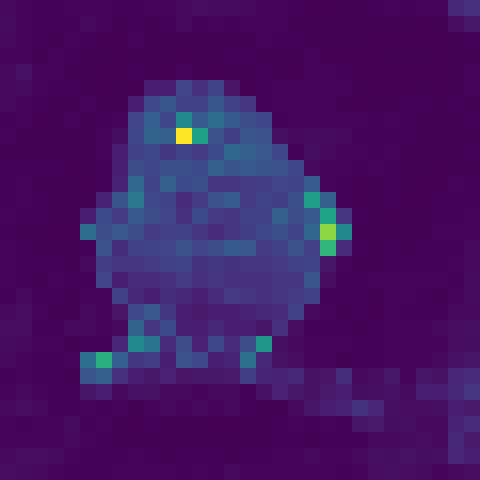}
        \end{subfigure}
    \end{subfigure}

        \begin{subfigure}[t]{\linewidth}
    \centering
    
        \begin{subfigure}[t]{0.085\linewidth}
        \includegraphics[width=\textwidth]{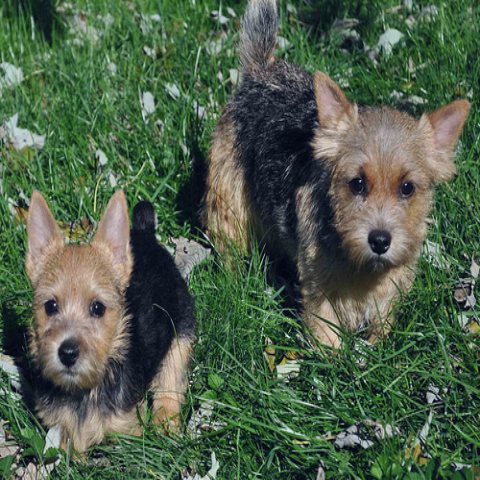}
        \end{subfigure}
        \begin{subfigure}[t]{0.085\linewidth}
        \includegraphics[width=\textwidth]{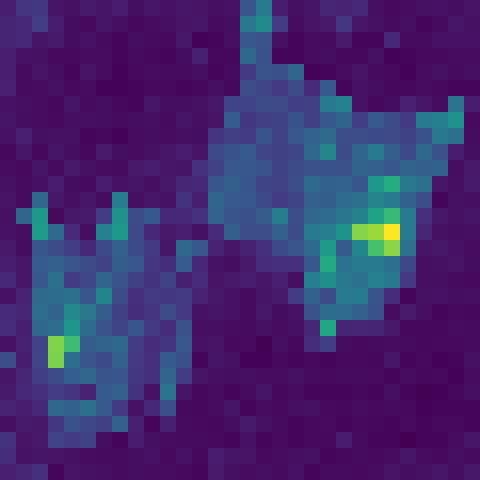}
        \end{subfigure}
        \hspace{0.15cm}
        \begin{subfigure}[t]{0.085\linewidth}
        \includegraphics[width=\textwidth]{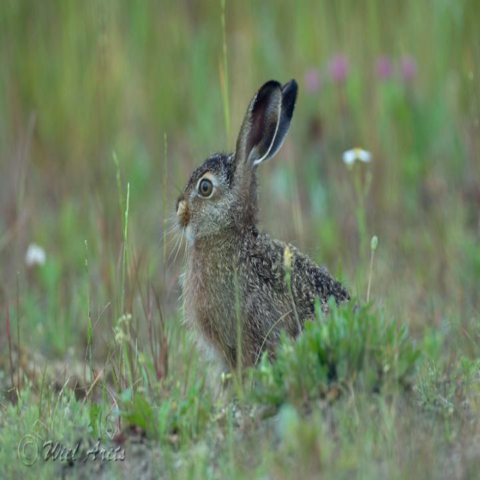}
        \end{subfigure}
        \begin{subfigure}[t]{0.085\linewidth}
        \includegraphics[width=\textwidth]{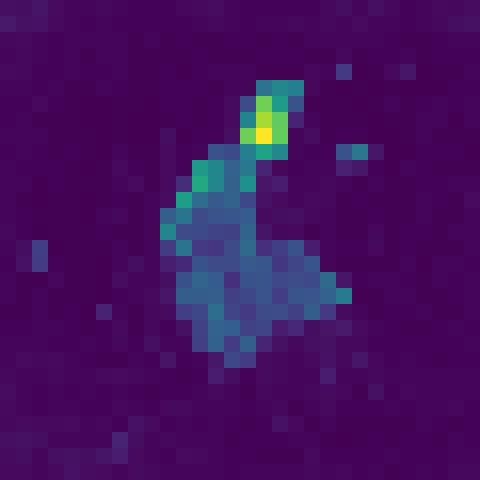}
        \end{subfigure}
        \hspace{0.15cm}
        \begin{subfigure}[t]{0.085\linewidth}
        \includegraphics[width=\textwidth]{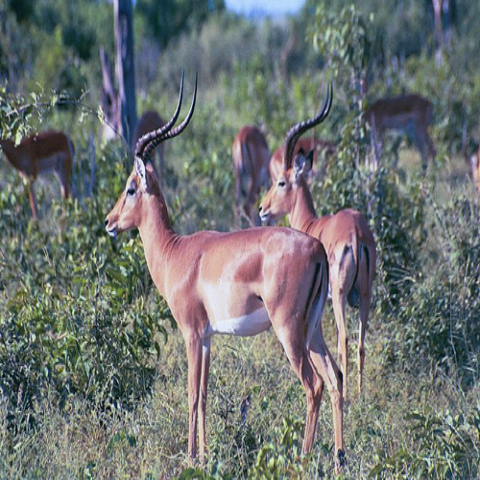}
        \end{subfigure}
        \begin{subfigure}[t]{0.085\linewidth}
        \includegraphics[width=\textwidth]{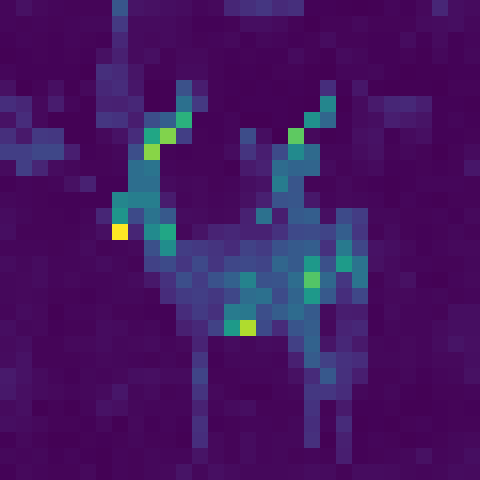}
        \end{subfigure}
        \hspace{0.15cm}
        \begin{subfigure}[t]{0.085\linewidth}
        \includegraphics[width=\textwidth]{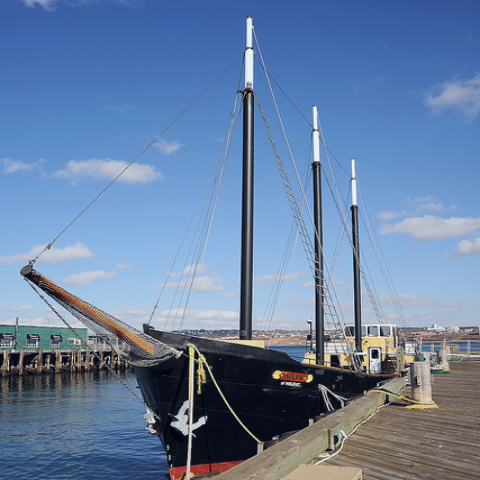}
        \end{subfigure}
        \begin{subfigure}[t]{0.085\linewidth}
        \includegraphics[width=\textwidth]{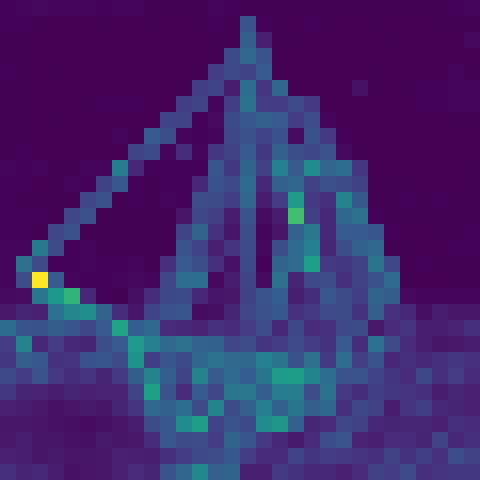}
        \end{subfigure}
        \hspace{0.15cm}
        \begin{subfigure}[t]{0.085\linewidth}
        \includegraphics[width=\textwidth]{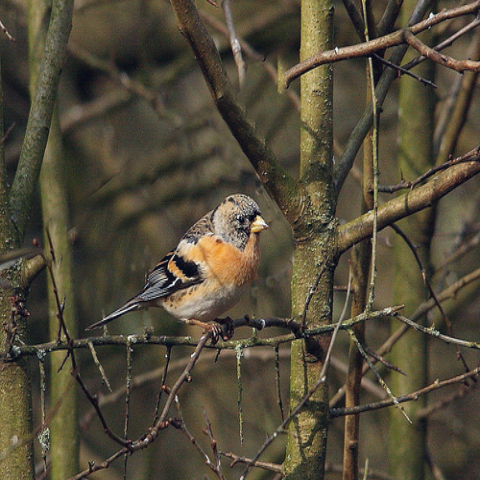}
        \end{subfigure}
        \begin{subfigure}[t]{0.085\linewidth}
        \includegraphics[width=\textwidth]{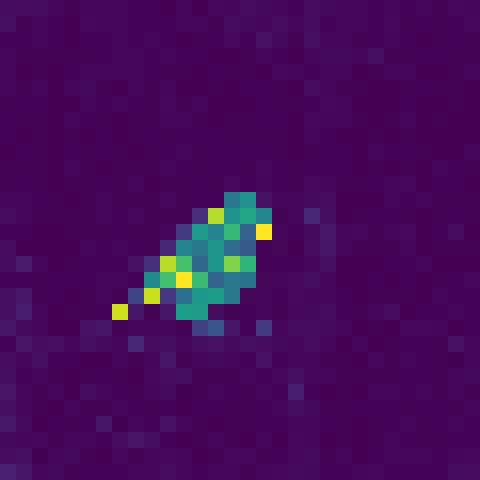}
        \end{subfigure}
    \end{subfigure}
    \caption{Attention of the class token [CLS] after the pre-training stage of SiT in an unsupervised fashion on ImageNet dataset employing ViT-S/16 vision transformer.}
    \label{fig:vis}

\end{figure*}